\documentclass[journal]{IEEEtran}
\usepackage{amsmath,amsfonts}
\usepackage{algorithmic}
\usepackage{algorithm}
\usepackage{array}
\usepackage[caption=false,font=normalsize,labelfont=sf,textfont=sf]{subfig}
\usepackage{textcomp}
\usepackage{stfloats}
\usepackage{url}
\usepackage{verbatim}
\usepackage{graphicx}
\usepackage{xcolor}
\usepackage{cite}
\usepackage{booktabs}
\usepackage{multirow}

\begin{document}

\title{Globally Optimal Solution to the Generalized Relative Pose Estimation Problem using Affine Correspondences}

\author{
	\vskip 1em
	Zhenbao Yu, Banglei Guan*, Shunkun Liang, Zibin Liu, Yang Shang, and Qifeng Yu

\thanks{Zhenbao Yu  is with the College of Aerospace Science and Engineering, National University of Defense Technology, Changsha 410000, China, and also with the Global Navigation Satellite System Research Center, Wuhan University, Wuhan 430000, China. (zhenbaoyu@whu.edu.cn)}
\thanks{Banglei Guan, Shunkun Liang, Zibin Liu, Yang Shang, and Qifeng Yu are with the College of Aerospace Science and Engineering, National University of Defense Technology, Changsha 410000, China.(\{guanbanglei12, liangshunkun, liuzibin19, shangyang1977, yuqifeng\}@nudt.edu.cn)}}

\markboth{}%
{Shell \MakeLowercase{\textit{et al.}}: A Sample Article Using IEEEtran.cls for IEEE Journals}

\IEEEpubid{0000--0000/00\$00.00~\copyright~**** IEEE}

\maketitle

\begin{abstract}
Mobile devices equipped with a multi-camera system and an inertial measurement unit (IMU) are widely used nowadays, such as self-driving cars. The task of relative pose estimation using visual and inertial information has important applications in various fields. To improve the accuracy of relative pose estimation of multi-camera systems, we propose a globally optimal solver using affine correspondences to estimate the generalized relative pose with a known vertical direction. First, a cost function about the relative rotation angle is established after decoupling the rotation matrix and translation vector, which minimizes the algebraic error of geometric constraints from affine correspondences. Then, the global optimization problem is converted into two polynomials with two unknowns based on the characteristic equation and its first derivative is zero. Finally, the relative rotation angle can be solved using the polynomial eigenvalue solver, and the translation vector can be obtained from the eigenvector. Besides, a new linear solution is proposed when the relative rotation is small. The proposed solver is evaluated on synthetic data and real-world datasets. The experiment results demonstrate that our method outperforms comparable state-of-the-art methods in accuracy.
 \end{abstract}


\section{Introduction}
\begin{figure}[t]
	\centering 
	\includegraphics[width=0.9\linewidth]{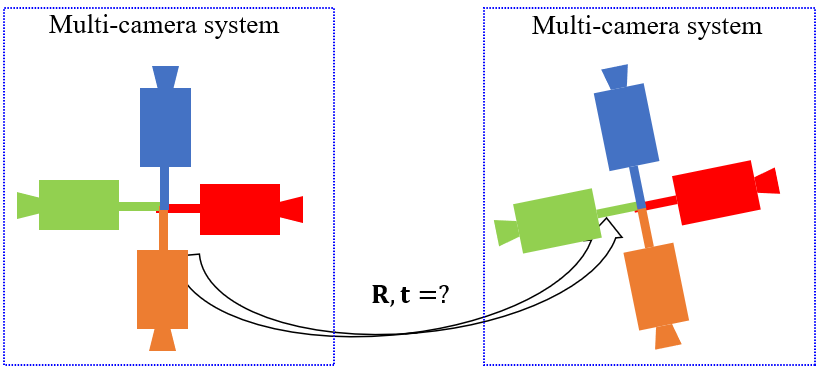}
	\caption{Relative pose estimation of a multi-camera system. The rotation matrix and translation vector between two views are ${\bf{R}}$ and ${\bf{t}}$, respectively.} 
 \label{fig1}
\end{figure}

\IEEEPARstart{O}{ne} of the most fundamental problems in geometric vision is to calculate the relative pose of two views. It plays an important role in visual localization (VO)~\cite{VO}, simultaneous localization and mapping (SLAM)~\cite{SLAM1,SLAM2,SLAM3}, and structure-from-motion (SfM)~\cite{SMF1,SMF2,SMF3}. To improve the accuracy and efficiency of relative pose estimation, a large number of algorithms have been developed. Since the multi-camera systems have the advantage of a large field of view, which can obtain more environmental information, they have been extensively used in self-driving cars, vehicle robots, and micro aerial vehicles. As shown in Fig.\ref{fig1}, this paper focuses on the generalized relative pose estimation of multi-camera systems.

The primary distinction between the single-camera system and the multi-camera system lies in the absence of a singular projection center in the latter. A pinhole or perspective camera model is used to describe a single-camera system, while a multi-camera system is represented by a generalized camera model (GCM). In multi-camera systems, light rays from different 3D points do not converge at a common optical center. The generalized essential matrix (GEM) and Pl{\"u}cker line vector are proposed in~\cite{plucker}. The translation obtained by a single camera system is without scale information. However, the translation extracted in the GEM includes scale information. The degree of freedom of the generalized relative pose is six between two views of multi-camera systems. Therefore, estimating the generalized relative pose requires at least six point correspondences in a multi-camera system.

A large number of methods have been proposed to estimate the generalized relative pose of multi-camera systems~\cite{6pt_Stewenius,4pt_lee,4pt_liu,4pt_Sweeney,17pt_li}. The first minimal solver using six point correspondences is derived and up to 64 solutions can be obtained\cite{6pt_Stewenius}. The linear solver with 17 point correspondences is proposed in~\cite{17pt_li}. This method is easy to implement when the number of corresponding points is greater than 17. Furthermore, some scholars utilize affine correspondences to solve the generalized relative pose of multi-camera systems~\cite{guan_6DOF,ICCV,IJCV}. These works prove the superiority of affine correspondences in generalized relative pose estimation problems.
\IEEEpubidadjcol

There are also some methods of using non-minimum samples to estimate the generalized relative pose of the multi-camera systems. Kneip and Li propose a solver using non-minimum sample points~\cite{8pt_Kneip}, but this method is greatly affected by the initial value. At present, global optimization methods utilize point correspondences in the generalized relative pose estimation of multi-camera systems~\cite{8pt_Kneip,zhao2020certifiably,Ding_mul}. As far as we know, there are currently no global optimal methods to estimate the generalized relative pose using affine correspondences.

To improve the efficiency of generalized relative pose estimation, some sensors, such as IMU, are usually attached to the multi-camera systems. In this case, the vertical direction information is provided by the IMU. Therefore, the degree of freedom of generalized relative pose is reduced to 4, including 1 for rotation and 3 for translation. Minimal solutions and non-minimal solutions with a known  vertical direction are proposed in~\cite{4pt_lee,4pt_liu,4pt_Sweeney,Ding_mul,ICCV,banglei2020relative}. 

In this paper, we mainly focus on globally optimal generalized relative pose estimation of the multi-camera systems using affine correspondences. Nowadays, mobile devices equipped with a multi-camera system and an IMU are widely used. We assume that the IMU provides pitch and roll angle information for the multi-camera systems, which means that the vertical direction can be obtained.

There are three differences between our method and previous works: (1) We are the first to propose a global optimization solver from N-affine correspondences instead of N-point correspondences~\cite{Ding_mul}. The existing method of calculating the relative pose of multiple camera systems by affine correspondence uses the minimum sample~\cite{ICCV,banglei2020relative}. The method proposed in this paper fully utilizes affine correspondences. Experimental results indicate that our method performs well in multi-camera relative pose estimation. (2) We derive the decoupling process of the rotation matrix and translation vector in the affine transformation constraint for multi-camera systems when the vertical direction is known. Besides, we provide expressions for decoupling the rotation matrix and translation vector in affine transformation constraint, which laid the foundation for computing the relative pose using affine correspondences for future utilization. However, previous methods only provided expressions for decoupling the rotation matrix and translation vector in generalized epipolar constraints~\cite{Ding_mul}. (3) We derive a polynomial eigenvalue solver applicable to the cost function established using affine correspondences and provided its expression. The solver can efficiently solve for the relative pose. This polynomial eigenvalue solver offers a reference for solving the relative pose using the affine transformation constraint in the future. However, previous methods provided a solver applicable to the cost function established using point correspondences~\cite{Ding_mul}.

The major contributions of this paper are three folds:

$ \bullet $ A globally optimal solver with N-affine correspondences is proposed for the multi-camera system when the vertical direction is known. By using the geometric constraints of affine correspondences, the cost function about the relative rotation angle is established based on minimizing algebraic error using least squares estimation.

$ \bullet $ The cost function is converted to finding the minimum eigenvalue of the matrix containing the relative rotation angle. Based on the characteristic equation method and its first derivative is zero, two independent polynomial equations with two unknowns are derived. Then, the eigenvalue polynomial solver is used to solve the rotation angle parameter. And the translation vector is obtained from the eigenvector.

$ \bullet $ We provide decoupling results for the rotation matrix and translation vector in the affine transformation constraint in multi-camera systems with a known vertical direction. Besides, a new linear solution using first-order approximation is proposed with N-affine correspondences when the relative rotation is small.

The rest of this paper is organized as follows. In Section 2, the related work is introduced. The generalized epipolar constraint and affine transformation constraint are introduced in Section 3. The cost function is established in Section 4. The solver for solving the cost function is proposed in Section 5. We propose a linear solver using first-order approximation in Section 6. The experimental results are shown in Section 7. A summary discussion is provided in the final section.

\section{Related Work}
The relative pose estimation of the multi-camera systems has received extensive attention in academia and industry. Our work focuses on global optimization algorithms using non-minimal samples. A solver using second order cone programming (SOCP) is proposed in~\cite{SOCP}. This method simplifies the motion estimation problem to estimate a triangulation problem and utilizes SOCP to find the optimal solution. However, this method does not provide a unified framework for multi-camera pose estimation~\cite{bound_bound}. A globally optimal solver using the branch-and-bound method is proposed, applying the rotation space search technique proposed by Hartley and Kakl to find the optimal solution~\cite{bound_bound}. Besides, this method requires other methods to provide the initial relative pose. The relative pose solver of the multi-camera systems can be estimated by finding the smallest eigenvalue proposed in~\cite{8pt_Kneip}. However, this method sometimes provides a locally optimal solution rather than a globally optimal solution. A global optimization framework is proposed in~\cite{5_Ding}. 

Estimating the essential matrix by algebraic error minimization has been extensively studied in relative pose estimation and can be formulated as a polynomial problem. A method to solve non-convex optimization problems with polynomials using the convex relaxations method is proposed in~\cite{QCQP}. Building on this foundation, a convex optimization approach for relative pose estimation of the multi-camera system is proposed in~\cite{zhao2020certifiably}. This method transforms the minimization of the sum of squared residuals, constructed by generalized epipolar constraints, to a quadratically constrained quadratic program (QCQP). Then, the global optimal GEM is obtained by a semidefinite relaxation. Additionally, a sufficient and necessary condition for global optimality from the relaxed problems is provided. A new globally optimal solver for the multi-camera system with a known vertical direction is proposed in~\cite{Ding_mul}. 
Polynomial eigenvalue solvers are widely utilized in computer vision~\cite{bujnak20093d,fitzgibbon2001simultaneous,ding2020efficient,polynomial}. For instance, these solvers are employed to determine the relative positions of cameras and a single unknown focal length from 6 point correspondences~\cite{bujnak20093d}. This solver can generate 9 solutions. The polynomial eigenvalue solver is applied in multiple-view geometry and lens distortion~\cite{fitzgibbon2001simultaneous}. For a calibrated camera, the relative pose of the camera can be estimated from 3 point correspondences in camera-IMU systems. In this case, the polynomial eigenvalue solver is applied to calculate the relative pose~\cite{ding2020efficient}. A detailed introduction to the eigenvalue polynomial solver is provided in~\cite{polynomial}. The solution methods for common problems in relative pose estimation are also provided.

Methods based on deep learning have also been widely applied in relative pose estimation. Techniques for learning frame-to-frame motion fields using deep neural networks are proposed~\cite{Dosovitskiy2015FlowNetLO,mayer2016large}. SfM-Net is introduced in~\cite{vijayanarasimhan1704sfm}. This method decomposes frame-by-frame pixel motion based on scene and object depth and camera motion, as well as three-dimensional object rotation and translation. The literature review and summarize the SFM while proposing a new deep learning-based two-view SFM framework~\cite{wang2021deep}. A method that combines traditional geometric approaches with deep learning methods is proposed~\cite{zhuang2021fusing}. An end-to-end NFlowNet network is introduced to estimate camera relative pose~\cite{parameshwara2022diffposenet}. To improve the robustness and generalization of the end-to-end two-view SfM network, the two-view SfM problem is formulated as maximum likelihood estimation and solved using the proposed DeepMLE framework~\cite{xiao2022deepmle}.

Recently, affine correspondence has garnered attention among scholars. The explicit relationship between the essential matrix and local affine transformation is derived in~\cite{AC_EX}. Four solvers using affine correspondence for relative pose estimation are proposed in~\cite{guan2020minimal,guan2021relative}. However, these solvers are limited to a single-camera system. In the multi-camera system, a series of solvers for relative pose estimation from two affine correspondences is proposed in~\cite{zhao_jiaocha}. A new constraint between the generalized essential matrix and local affine transformation is derived in~\cite{IJCV}. An affine correspondence is sufficient to estimate the relative pose where the motion is planar. Besides, two affine correspondences can estimate essential matrices in the case of random motion.

Currently, relative pose solutions using affine correspondences are based on a minimal number of affine correspondences in the multi-camera system~\cite{ICCV,banglei2020relative}. Experimental results demonstrate that the accuracy of relative pose computed using affine correspondences is superior to that computed using point correspondences. A globally optimal solver using point correspondence to estimate generalized relative pose with vertical direction is proposed in~\cite{Ding_mul}. The global optimization solver proposed in our paper is based on affine correspondences with a non-minimal number of samples. We establish a cost function based on minimizing the algebraic error of affine transformation constraint and generalized epipolar constraint. The cost function is established solely using generalized epipolar constraint in~\cite{Ding_mul}. In the actual application process, there are sometimes situations where the rotation angle between two consecutive frames is very small. Therefore, it is very meaningful to propose a new first-order approximate linear solution with N-affine correspondences.

\begin{figure}[ttp]
	\centering 
	\includegraphics[width=0.9\linewidth]{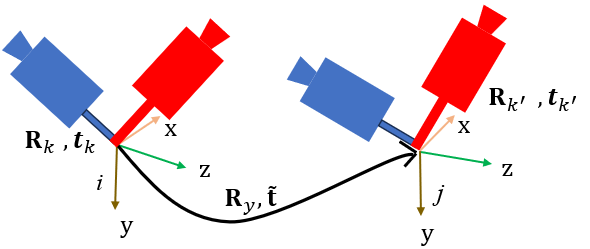}
	\caption{The rotation matrix and translation vector of the $k$-th camera are ${{\bf{R}}_{k}}$ and ${{\bf{t}}_{k}}$. The rotation matrix and translation vector of the $k'$-th camera are ${{\bf{R}}_{k^{'}}}$ and ${{\bf{t}}_{k^{'}}}$. The rotation matrix and translation vector between the aligned views at moment $i$ and $j$ are ${{\bf{R}}_y}$ and ${\bf{\tilde t}}$.}
 \label{fig2}
\end{figure}

\section{Geometric Constraints}
The multi-camera systems consist of multiple individual cameras fixed to a single rigid body. An affine correspondence consists of a point correspondence and a local affine transformation. Local affine transformations represent the warp between the infinitely close vicinities of corresponding point pairs. A point correspondence can provide constraint equations, and a local affine transformation can provide two constraints, so an affine correspondence can provide three constraint equations. Compared to point correspondences, affine correspondences provide more constraint equations. Denote an affine correspondence $({{\bf{x}}_{ki}},{\bf{x}}_{k^{'}j},{{\bf{R}}_{k}},{\bf{R}}_{k^{'}},{{\bf{t}}_{k}},{\bf{t}}_{k^{'}},{{{\bf{A}}}_{ij}})$, where ${{\bf{x}}_{ki}}$ is the normalized homogeneous coordinate of the $k$-th camera capture image at moment ${i}$, and ${{\bf{x}}_{k^{'}j}}$ is the normalized homogeneous coordinate of the $k'$-th camera capture image at moment ${j}$. The rotation matrix and translation vector of the $k$-th camera are ${{\bf{R}}_{k}}$ and ${{\bf{t}}_{k}}$. The rotation matrix and translation vector of the $k'$-th camera are ${{\bf{R}}_{k^{'}}}$ and ${{\bf{t}}_{k^{'}}}$ (Fig.\ref{fig2}). ${\bf{A}}_{ij}$ represents local affine transformation between points ${{\bf{x}}_{ki}}$ and ${\bf{x}}_{k^{'}j}$. If the affine correspondence is obtained by the same camera at the moment ${i}$ and ${j}$, then ${{\bf{R}}_{k}}={{\bf{R}}_{k^{'}}}$ and ${{\bf{t}}_{k}}={{\bf{t}}_{k^{'}}}$. This paper focuses on relative pose estimation when the IMU is coupled with the multi-camera system. The pitch and roll angles of the reference frame can be obtained from the IMU. The rotation matrices provided by the IMU at moment ${i}$ and ${j}$ are ${\bf{R}}_{{\rm{imu}}}$ and ${\bf{R}}_{{\rm{imu}}}^{'}$,  respectively. ${{\bf{R}}_{{\rm{imu}}}}$ can be expressed as :
\begin{align}
	{\bf {R_{{\rm{imu}}}} }={{\bf{R}}_x}{{\bf{R}}_z},
   \label{eq1}
\end{align}
where ${{\bf{R}}_x}$ and ${\bf{R}}_z$ can be written as:
\begin{equation}
	\begin{aligned}
     	{{\bf{R}}_x} = \left[ {\begin{array}{*{20}{c}}
		1&0&0\\
		0&{{\rm{cos}}\left( {{\theta _x}} \right)}&{-{\rm{sin}}\left( {{\theta _x}} \right)}\\
		0&{  {\rm{sin}}\left( {{\theta _x}} \right)}&{{\rm{cos}}\left( {{\theta _x}} \right)}
       \end{array}} \right],
	\end{aligned}
	\label{eq2}
\end{equation}
\begin{equation}
	\begin{aligned}
      {{\bf{R}}_z} = \left[ {\begin{array}{*{20}{c}}
		{{\rm{cos}}\left( {{\theta _z}} \right)}&{-{\rm{sin}}\left( {{\theta _z}} \right)}&0\\
		{  {\rm{sin}}\left( {{\theta _z}} \right)}&{{\rm{cos}}\left( {{\theta _z}} \right)}&0\\
		0&0&1
      \end{array}} \right],
	\end{aligned}
	\label{eq3}
\end{equation}
where ${\theta _x}$ is the roll angle and ${\theta _z}$ is the pitch angle. In this case,  ${{\bf{R}}_y}$ and  ${\bf{\tilde t}}$ represent the rotation and translation between the aligned views at moment ${i}$ and ${j}$.
\begin{equation}
	\begin{aligned}
      {{\bf{R}}_y} = \left[ {\begin{array}{*{20}{c}}
		{{\rm{cos}}\left( {{\theta _y}} \right)}&0&{{\rm{sin}}\left( {{\theta _y}} \right)}\\
		0&1&0\\
		{-{\rm{sin}}\left( {{\theta _y}} \right)}&0&{{\rm{cos}}\left( {{\theta _y}} \right)}
      \end{array}} \right],
	\end{aligned}
	\label{eq4}
\end{equation}
\begin{equation}
	\begin{aligned}
  {\bf{\tilde t}} = \left[ {\begin{array}{*{20}{c}}
{{{\tilde t}_x}}&{{{\tilde t}_y}}&{{{\tilde t}_z}}
\end{array}} \right]^T,
	\end{aligned}
	\label{eq5}
\end{equation}
where $\theta_y$ is the rotation angle between the aligned views. The rotation matrix ${{\bf{R}}_y}$ using Cayley parameterization can be written as:
\begin{equation}
	\begin{aligned}
       {{\bf{R}}_y} = \frac{1}{{1 + {s^2}}}\left[ {\begin{array}{*{20}{c}}
	     	{1 - {s^2}}&0&{2s}\\
	        	0&1&0\\
	    	{ - 2s}&0&{1 - {s^2}}
      \end{array}} \right],
	\end{aligned}
	\label{6}
\end{equation}
where $s$ is represented as $\tan(\theta_y/2)$.

\subsection{Generalized epipolar constraint}
In this section, we first briefly describe the concept of generalized epipolar constraint for a multi-camera system. A Pl{\"u}cker vector is often used when estimating the relative pose of a multi-camera system~\cite{plucker}. A Pl{\"u}cker vector is ${6 \times 1}$ vector, including the direction vector of the ray (the first three entries) and the moment of the corresponding line (the latter three entries). The generalized epipolar constraint is written as:
\begin{equation}
	\begin{aligned}
{\bf{I}}{_{k'j}^T}\left[ {\begin{array}{*{20}{c}}
{\bf{E}}&{\bf{R}}\\
{\bf{R}}&0
\end{array}} \right]{{\bf{I}}_{ki}} = 0,
	\end{aligned}
	\label{7}
\end{equation}
where ${{\bf{I}}_{ki}}$ and ${\bf{I}}_{k'j}$ denote a pair of corresponding Pl{\"u}cker-vectors at moment ${i}$ and ${j}$. The Pl{\"u}cker vector is written as:
\begin{equation}
	\begin{aligned}
{{\bf{I}}_{ki}} = \left( {\begin{array}{*{20}{c}}
{{{\bf{f}}_{ki}}}\\
{{{\bf{t}}_{k}} \times {{\bf{f}}_{ki}}}
\end{array}} 
\right),
\quad
{{\bf{I}}_{k'j}} = \left( {\begin{array}{*{20}{c}}
{{{\bf{f}}_{k'j}}}\\
{{{\bf{t}}_{k'}} \times {{\bf{f}}_{k'j}}}
\end{array}} \right),
\end{aligned}
	\label{8}
\end{equation}
where ${\bf{f}}_{ki}$ and ${\bf{f}}_{k'j}$  are written as:
\begin{equation}
	\begin{aligned}
{{\bf{f}}_{ki}} = \frac{{({{\bf{R}}_{k}}{{\bf{x}}_{ki}})}}{{\left\| {{{\bf{R}}_{k}}{{\bf{x}}_{ki}}} \right\|}},
\qquad{}
{{\bf{f}}_{k'j}} = \frac{{({{\bf{R}}_{k'}}{{\bf{x}}_{k'j}})}}{{\left\| {{{\bf{R}}_{k'}}{{\bf{x}}_{k'j}}} \right\|}}.
	\end{aligned}
 \label{9}
\end{equation}
The essential matrix $\bf{E}$ is written as:
\begin{align}
    \bf E = {\left[ t \right]_ \times }R,
    \label{10}
 \end{align}
where ${{\bf{R}}}$ and  ${\bf{t}}$ represent the rotation and translation between the unaligned views at moment ${i}$ and ${j}$, respectively. Combined with Eqs.~\eqref{eq1} and~\eqref{eq5}, ${\bf{R}}$ and ${\bf{t}}$ can be expressed as:
\begin{equation}
	\begin{aligned}
{\bf{R}} = {({\bf{R}}_{{\rm{imu}}}')^T}{{\bf{R}}_y}{{\bf{R}}_{{\rm{imu}}}}
	\end{aligned},
	\label{11}
\end{equation}
\begin{equation}
	\begin{aligned}
{\bf{t}} = {({\bf{R}}_{{\rm{imu}}}')^T}{\bf{\tilde t}}
	\end{aligned}.
	\label{12}
\end{equation}

We substitute Eq.~\eqref{11} and Eq.~\eqref{12} into Eq.~\eqref{7}
\begin{equation}
	\begin{aligned}
{\left( {\left[ {\begin{array}{*{20}{c}}
{{\bf{R}}_{{\rm{imu}}}'}&0\\
0&{{\bf{R}}_{{\rm{imu}}}^{'}}
\end{array}} \right]{\bf{I}}_{k'j}} \right)}^T \cdot
\left[ {\begin{array}{*{20}{c}}
{{{\left[ {\bf{\tilde t}} \right]}_ \times }{{\bf{R}}_y}}&{{{\bf{R}}_y}}\\
{{{\bf{R}}_y}}&0
\end{array}} \right] \cdot\\
\left( {\left[ {\begin{array}{*{20}{c}}
{{{\bf{R}}_{{\rm{imu}}}}}&0\\
0&{{{\bf{R}}_{{\rm{imu}}}}}
\end{array}} \right]{{\bf{I}}_{ki}}} \right) = 0 
\end{aligned} \textcolor{red}{.}
\label{13}
\end{equation}

By substituting Eq.~\eqref{8} into  Eq.~\eqref{13}, we can get
\begin{equation}
	\begin{aligned}
{\left( {\begin{array}{*{20}{c}}
{{{\bf{R}}_{{\rm{imu}}}'}{{\bf{f}}_{{k'}j}}}\\
{{{\bf{R}}_{{\rm{imu}}}'}({{\bf{t}}_{{k'}}} \times {{\bf{f}}_{{k'}j}})}
\end{array}} \right)^T}\cdot\left[ {\begin{array}{*{20}{c}}
{{{\left[ {{\bf{\tilde t}}} \right]}_ \times }{{\bf{R}}_y}}&{{{\bf{R}}_y}}\\
{{{\bf{R}}_y}}&0
\end{array}} \right]\cdot \\
\left( {\begin{array}{*{20}{c}}
{{{\bf{R}}_{{\rm{imu}}}}{{\bf{f}}_{ki}}}\\
{{{\bf{R}}_{{\rm{imu}}}}({{\bf{t}}_{k}} \times {{\bf{f}}_{ki}})}
\end{array}} \right) = 0 
\end{aligned}.
\label{14}
\end{equation}

Eq.~\eqref{14} can be simplified as:
\begin{equation}
	\begin{aligned}
  {\left( {\begin{array}{*{20}{c}}
{{({\bf{R}}_{{\rm{imu}}}'}{{\bf{f}}_{{k'}j}}) \times {{\bf{R}}_y}({{\bf{R}}_{{\rm{imu}}}}{{\bf{f}}_{ki}})}\\
{{O_1}}
\end{array}} \right)^T}{\widehat {\bf{t}}} = 0
\end{aligned},
\label{15}
\end{equation}
where $\widehat {\bf{t}}$=${[\begin{array}{*{20}{c}}{{\bf{\tilde t}}}&1\end{array}]^T}$ and ${{O_1}}$ can be written as: 
\begin{equation}
	\begin{aligned}
{O_1} = {\bf{f}}_{{k'}j}^T({\left[ {{{\bf{t}}_{k'}}} \right]_ \times }{({\bf{R}}_{{\rm{imu}}}')^T}{{\bf{R}}_y}{{\bf{R}}_{{\rm{imu}}}} -\\ {({\bf{R}}_{{\rm{imu}}}')^T}{{\bf{R}}_y}{{\bf{R}}_{{\rm{imu}}}}{\left[ {{{\bf{t}}_{k}}} \right]_ \times }){{\bf{f}}_{ki}}
\end{aligned}.
\label{16}
\end{equation}

\subsection{Affine transformation constraint} 
We denote the rotation and translation the ${k}$-th camera at moment ${i}$ to the ${k'}$-th camera at moment ${j}$ as ${{\bf{R}}_{ij}}$ and ${{\bf{t}}_{ij}}$ , which can be written as: 
\begin{equation}
\setlength{\arraycolsep}{1.8pt}
	\begin{aligned}
&\left[ {\begin{array}{*{20}{c}}
{{{\bf{R}}_{ij}}}&{{{\bf{t}}_{ij}}}\\
0&1
\end{array}} \right] = {\left[ {\begin{array}{*{20}{c}}
{{{\bf{R}}_{k'}}}&{{{\bf{t}}_{k'}}}\\
0&1
\end{array}} \right]}^{-1}\left[ {\begin{array}{*{20}{c}}
{\bf{R}}&{\bf{t}}\\
0&1
\end{array}} \right]\left[ {\begin{array}{*{20}{c}}
{{{\bf{R}}_{k}}}&{{{\bf{t}}_{k}}}\\
0&1
\end{array}} \right]\\& = \left[ {\begin{array}{*{20}{c}}
{{\bf{R}}_{k'}^T{\bf{R}}{{\bf{R}}_{k}}}&{{\bf{R}}_{k'}^T({\bf{R}}{{\bf{t}}_{k}} + {\bf{t}} - {{\bf{t}}_{k'}})}\\
0&1
\end{array}} \right]
	\end{aligned}.
	\label{17}
\end{equation}

According to Eq.~\eqref{17}, we can easily get the essential matrix between the ${k}$-th camera at moment ${i}$ and the ${k'}$-th camera at moment ${j}$, which can be written as: 
\begin{equation}
	\begin{aligned}
{{\bf{E}}_{ij}} &= {\left[ {{{\bf{t}}_{ij}}} \right]_ \times }{{\bf{R}}_{ij}} \\&= {\left[ {{\bf{R}}_{k'}^T({\bf{R}}{{\bf{t}}_{k}} + {\bf{t}} - {\bf{t}}_{k'}^T)} \right]_ \times }{\bf{R}}_{k'}^T{\bf{R}}{{\bf{R}}_{k}} \\&= {\bf{R}}_{k'}^T{\left[ {({\bf{R}}{{\bf{t}}_{k}} + {\bf{t}} - {\bf{t}}_{k'}^T)} \right]_ \times }{\bf{R}}{{\bf{R}}_{k}} \\&= {\bf{R}}_{k'}^T({\bf{R}}{\left[ {{{\bf{t}}_{k}}} \right]_ \times } + {\left[ {{\bf{t}} - {{\bf{t}}_{k'}}} \right]_ \times }{\bf{R}}){{\bf{R}}_{k}}
	\end{aligned}.
 \label{18}
\end{equation}

To simplify Eq.~\eqref{18}, a property that ${\left[ {{\bf{Rt}}} \right]_ \times }{\bf{R}} = {\bf{R}}{\left[ {\bf{t}} \right]_ \times },\forall  \in {\rm{SO}}(3)$ is exploited. The above equation can be represented as: 
\begin{equation}
	\begin{aligned}
{{\bf{E}}_{ij}}{\rm{ }}& =  {\bf{R}}_{k'}^T({({\bf{R}}_{_{{\rm{imu}}}}')^T}{{\bf{R}}_y}{{\bf{R}}_{{\rm{imu}}}}{\left[ {{{\bf{t}}_{k}}} \right]_ \times } +\\&
{\left[ {{{({\bf{R}}_{_{{\rm{imu}}}}')}^T}{\bf{\tilde t}} - {{\bf{t}}_{k'}}} \right]_ \times }{({\bf{R}}_{_{{\rm{imu}}}}')^T}{{\bf{R}}_y}{{\bf{R}}_{{\rm{imu}}}}){{\bf{R}}_{k}}
	\end{aligned}.
 \label{19}
\end{equation}

The relationship between the essential ${{{\bf{E}}_{ij}}}$ and local affine transformation ${\bf{A}}_{ij}$~\cite{ICCV} is as follows
\begin{equation}
	\begin{aligned}
{({\bf{E}}_{_{ij}}^T{{\bf{x}}_{{k'}j}})_{(1:2)}} =  - {({{\bf{\hat A}}_{ij}}{{\bf{E}}_{ij}}{{\bf{x}}_{ki}})_{(1:2)}}
	\end{aligned},
 \label{20}
\end{equation}
\begin{equation}
{{\bf{\hat A}}_{ij}} = \left[ {\begin{array}{*{20}{c}}
{{\bf{A}}_{ij}}&{\bf{0}}\\
{\bf{0}}&1
\end{array}} \right],
\label{21}
\end{equation}
where the ${(1:2)}$ represents the first two lines of the determinant. Based on Eqs.~\eqref{19},~\eqref{20}, and~\eqref{21}, we can get 
\begin{small} 
\begin{equation}
	\begin{aligned}
\left[ {\begin{array}{*{20}{c}}
{{\bf{R}}_{_{{\rm{imu}}}}'{{\bf{R}}_{k'}}{{\bf{x}}_{k'j}} \times {{\bf{R}}_y}{{\bf{c}}_1} + {{\bf{c}}_3} \times {{\bf{R}}_y}{{\bf{R}}_{{\rm{imu}}}}{{\bf{R}}_{k}}}{{\bf{x}}_{ki}}&{{O_2}}\\
{{\bf{R}}_{_{{\rm{imu}}}}'{{\bf{R}}_{k'}}{{\bf{x}}_{k'j}} \times {{\bf{R}}_y}{{\bf{c}}_2} + {{\bf{c}}_4} \times {{\bf{R}}_y}{{\bf{R}}_{{\rm{imu}}}}{{\bf{R}}_{k}}}{{\bf{x}}_{ki}}&{{O_2}}
\end{array}} \right]{\bf{\hat t}} = 0,
\end{aligned}
 \label{22}
\end{equation}
\end{small}
where ${{{\bf{c}}_1}}$ and ${{{\bf{c}}_2}}$ are the first and second columns of ${{{\bf{R}}_{{\rm{imu}}}}{{\bf{R}}_{k}}}$. ${{{\bf{c}}_3}}$ and ${{{\bf{c}}_4}}$ are the first and second rows of ${\bf{\hat A}}_{_{ij}}^T{({\bf{R}}_{\rm{imu}}'{{\bf{R}}_{k'}})^T}$. 
\section{establishing the cost function}
An affine correspondence $({{\bf{x}}_{ki}},{\bf{x}}_{k^{'}j},{{\bf{R}}_{k}},{\bf{R}}_{k^{'}},{{\bf{t}}_{k}},{\bf{t}}_{k^{'}},\\{{{\bf{A}}}_{ij}})$
can provide three constraint equations
\begin{equation}
	\begin{aligned}
{\bf{m}}_i^T\widehat {\bf{t}} = 0
	\end{aligned},
 \label{24}
\end{equation}
where
\begin{small} 
\begin{equation}
	\begin{aligned}
{\bf{m}}_i^T \!\!=\!\! \left[ {\begin{array}{*{20}{c}}
\begin{array}{l}
\!\!\!\!\!\!{(({\bf{R}}_{_{{\rm{imu}}}}'{{\bf{f}}_{k'j}}) \times {{\bf{R}}_y}({{\bf{R}}_{{\rm{imu}}}}{{\bf{f}}_{ki}}))^T}\\
\!\!\!\!\!\!{({\bf{R}}_{_{{\rm{imu}}}}'{{\bf{R}}_{k'}}{{\bf{x}}_{k'j}} \times {{\bf{R}}_y}{{\bf{c}}_1} + {{\bf{c}}_3} \times {{\bf{R}}_y}{{\bf{R}}_{{\rm{imu}}}}{{\bf{R}}_k}{\bf{x}}_{ki})^T}
\end{array}&\begin{array}{l}
\!\!\!\!\!\!\!\!\!\!{O_1}\!\!\!\!\!\!\!\!\\
\!\!\!\!\!\!\!\!\!\!{O_2}\!\!\!\!\!\!\!\!
\end{array}\\
\!\!\!\!\!\!{{{({\bf{R}}_{_{{\rm{imu}}}}'{{\bf{R}}_{k'}}{{\bf{x}}_{k'j}} \times {{\bf{R}}_y}{{\bf{c}}_2} + {{\bf{c}}_4} \times {{\bf{R}}_y}{{\bf{R}}_{{\rm{imu}}}}{{\bf{R}}_k}{\bf{x}}_{ki})}^T}}&\!\!\!\!\!\!\!\!\!\!\!\!{{O_3}}\!\!\!\!\!\!\!\!
\end{array}} \right].
	\end{aligned}
 \label{25}
\end{equation}
\end{small}

If there are $N$ affine correspondences, we obtain 
\begin{equation}
	\begin{aligned}
{{\bf{M}}^T} \cdot { \widehat {\bf{t}}} = {({{\bf{m}}_1}...{{\bf{m}}_N})^T} \cdot { \widehat {\bf{t}}} = 0.
	\end{aligned}
         \label{26}
\end{equation}

In this paper, we focus on a globally optimal solution with N-affine correspondences (${\rm{N}>2}$). We establish the cost function based on least squares estimation, and the cost function can be described as: 
\begin{equation}
		\begin{aligned}
{\arg _{{{\bf{R}}_y},\widehat {\bf{t}}}}\min {{\bf{\widehat{t} }}^T}{\bf{C}}\widehat {\bf{t}},
		\end{aligned}
  \label{27}
\end{equation}
where ${\bf{C}} = {\bf{M}}{{\bf{M}}^T}$, which is a $4 \times 4$ matrix. Eq.~\eqref{27} represents the values of ${{\bf{R}}_{\rm{y}}}$  and ${{\bf{\widehat{t} }}^T}$ when ${{\bf{\widehat{t} }}^T}{\bf{C}}\widehat {\bf{t}}$ is minimized.

Suppose ${\lambda _{{\bf{C}},\min }}$ is the smallest eigenvalue of matrix $\bf{C}$, Eq.~\eqref{27} can be transformed into the following problems
\begin{equation}
	\begin{aligned}
        {{\bf{R}}_{\rm{y}}}{\rm{ = }}\arg {\min _{{{\bf{R}}_{\rm{y}}}}}{\lambda _{{\bf{C}},\min }}.
	\end{aligned}
   \label{29}
\end{equation}

Due to the fact that IMU provides a vertical direction, ${{\bf{R}}_{\rm{y}}}$ only contains one unknown variable ${s}$. This reduces it to a simpler one-dimensional space optimization problem rather than the full rotation in three-dimensional space. Therefore, this problem can be globally optimally solved by efficiently computing all stationary points. A stationary point is a point where the first derivative of a function is zero ($\frac{{{\rm{d}}{\lambda}}}{{{\rm{d}}s}} = 0$).

To solve for the eigenvalue ${\lambda}$, we can get
\begin{equation}
	\begin{aligned}
      \det ({\bf{C}} - \lambda {\bf{I}}) = {\lambda ^4} + {f_1}{\lambda ^3} + {f_2}{\lambda ^2} + {f_3}\lambda  + {f_4}
	\end{aligned},
 \label{30}
\end{equation}
where ${\bf{I}}$ represents a ${4\times4}$ identity matrix.
\begin{equation}
	\begin{aligned}
\left\{ \begin{array}{l}
{f_1} =  - trace({\bf{C}})\\
{f_2} = \frac{1}{2}({(trace({\bf{C}}))^2} - trace({{\bf{C}}^2}))\\
{f_3} =  - \frac{1}{6}{(trace({\bf{C}}))^3} + \frac{1}{2}trace({\bf{C}})*\\
\qquad trace({{\bf{C}}^2})- \frac{1}{3}trace({{\bf{C}}^3})\\
{f_4} = \det ({\bf{C}})
\end{array} \right.
	\end{aligned}.
 \label{31}
\end{equation}
Based on the characteristic equation method ($\det ({\bf{C}} - \lambda {\bf{I}})=0$), we can get
\begin{equation}
	\begin{aligned}
 {\lambda ^4} + {f_1}{\lambda ^3} + {f_2}{\lambda ^2} + {f_3}\lambda  + {f_4} = 0
	\end{aligned},
 \label{32}
\end{equation}
where ${f_1}, {f_2}, {f_3} ,{f_4}$ contain only the unknown ${s}$. For convenience of narration, we use ${\lambda}$ instead of ${\lambda _{{\bf{C}},\min }}$. If ${\lambda}$ is the smallest eigenvalue of ${\bf{C}}$, then derivative $\frac{{{\rm{d}}{\lambda}}}{{{\rm{d}}s}} = 0$. So we can get 
\begin{equation}
	\begin{aligned}
\frac{{d{f_1}}}{{ds}}{\lambda ^3} + \frac{{d{f_2}}}{{ds}}{\lambda ^2} + \frac{{d{f_3}}}{{ds}}\lambda  + \frac{{d{f_4}}}{{ds}} = 0
	\end{aligned}.
  \label{33}
\end{equation}

By defining that $\alpha  = 1 + {s^2}$, we can get
\begin{equation}
	\begin{aligned}
\begin{array}{l}
{\rm{{}}}{f_1} = \frac{{{g_1}}}{{{\alpha ^2}}}{\rm{  \qquad{}  }}{f_2} = \frac{{{g_2}}}{{{\alpha ^4}}}{\rm{  \quad{}   }}{f_3} = \frac{{{g_3}}}{{{\alpha ^6}}}{\rm{  \qquad{}   }}{f_4} = \frac{{{g_4}}}{{{\alpha ^8}}}\\
\frac{{d{f_1}}}{{ds}} = \frac{{{w_1}}}{{{\alpha ^3}}}{\rm{ \quad{} }}\frac{{d{f_2}}}{{ds}} = \frac{{{w_2}}}{{{\alpha ^5}}}{\rm{ \quad{} }}\frac{{d{f_3}}}{{ds}} = \frac{{{w_3}}}{{{\alpha ^7}}}{\rm{ \quad{} }}\frac{{d{f_4}}}{{ds}} = \frac{{{w_4}}}{{{\alpha ^9}}}
\end{array}
	\end{aligned},
 \label{34}
\end{equation}
where ${{g_1}}$, ${{g_2}}$, ${{g_3}}$, ${{g_4}}$, ${{w_1}}$, ${{w_2}}$, ${{w_3}}$, and ${{w_4}}$ are the polynomials of ${s}$. Table~\ref{tab:xishu} shows the highest degree of variables ${s}$. Multiplying ${\alpha ^8}$ to Eq.~\eqref{32} and Multiplying ${\alpha ^9}$ to Eq~\eqref{33} yields Eq.~\eqref{35}.

\begin{equation}
	\begin{aligned}
\left\{ \begin{array}{l}
{\beta ^4} + {\beta ^3}{g_1} + {\beta ^2}{g_2} + \beta {g_3} + {g_4} = 0\\
{\beta ^3}{w_1} + {\beta ^2}{w_2} + \beta {w_3} + {w_4} = 0
\end{array} \right.
	\end{aligned},
	\label{35}
\end{equation}
where $\beta  = {\alpha ^2}\lambda $. Eq.~\eqref{35} can be rewritten as:
\begin{equation}
	\begin{aligned}
\left[ {\begin{array}{*{20}{c}}
1&{{g_1}}&{{g_2}}&{{g_3}}&{{g_4}}\\
0&{{w_1}}&{{w_2}}&{{w_3}}&{{w_4}}
\end{array}} \right]\left[ {\begin{array}{*{20}{c}}
{{\beta ^4}}\\
{{\beta ^3}}\\
{{\beta ^2}}\\
\beta \\
1
\end{array}} \right] =\bf{0}
	\end{aligned}.
	\label{36}
\end{equation}

We obtain two polynomial equations with two unknowns ${\beta,{s}}$. The next section describes how to solve these polynomial equations accurately and quickly.
\section{Globally Optimal Solver}
It is easy to see that Eq.~\eqref{36} consists of two equations and five monomials (${\beta ^4}$, ${\beta ^3}$, ${\beta ^2}$, ${\beta }$, 1). We make the number of equations equal to the number of monomials by increasing the number of equations. The first equation is multiplied by {${\beta ^2}$, ${\beta }$} and the second equation is multiplied by {${\beta ^3}$, ${\beta ^2}$, ${\beta }$}. In this way, we can get five equations
\begin{equation}
	\begin{aligned}
\left\{ \begin{array}{l}
{\beta ^5} + {\beta ^4}{g_1} + {\beta ^3}{g_2} + {\beta ^2}{g_3} + \beta {g_4} = 0\\
{\beta ^6} + {\beta ^5}{g_1} + {\beta ^4}{g_2} + {\beta ^3}{g_3} + {\beta ^2}{g_4} = 0\\
{\beta ^4}{w_1} + {\beta ^3}{w_2} + {\beta ^2}{w_3} + \beta {w_4} = 0\\
{\beta ^5}{w_1} + {\beta ^4}{w_2} + {\beta ^3}{w_3} + {\beta ^2}{w_4} = 0\\
{\beta ^6}{w_1} + {\beta ^5}{w_2} + {\beta ^4}{w_3} + {\beta ^3}{w_4} = 0
\end{array} \right.
	\end{aligned}.
	\label{37}
\end{equation}

\begin{table}
 \centering
 \small
  \caption{Degree of ${g_i},{w_i}$ (${i=1,2,3,4}$).}
\begin{tabular}{ccccccccc} 
\hline
\toprule    
    & ${g_1}$ & ${g_2}$  & ${g_3}$& ${g_4}$ & ${w_1}$ & ${w_2}$  & ${w_3}$ & ${w_4}$  \\
\midrule 
Degree(s) & 4 & 8 & 12 & 16 & 4 & 8 & 12 & 16\\
\bottomrule  
\end{tabular}
 \label{tab:xishu}
\end{table}
Based on Eqs.~\eqref{35} and~\eqref{37}, we obtain seven equations with seven monomials, which can be expressed as:
\begin{small}
\begin{equation}
	\begin{aligned}
\underbrace {\left[ {\begin{array}{*{20}{c}}
0&0&1&{{g_1}}&{{g_2}}&{{g_3}}&{{g_4}}\\
0&1&{{g_1}}&{{g_2}}&{{g_3}}&{{g_4}}&0\\
1&{{g_1}}&{{g_2}}&{{g_3}}&{{g_4}}&0&0\\
0&0&0&{{w_1}}&{{w_2}}&{{w_3}}&{{w_4}}\\
0&0&{{w_1}}&{{w_2}}&{{w_3}}&{{w_4}}&0\\
0&{{w_1}}&{{w_2}}&{{w_3}}&{{w_4}}&0&0\\
{{w_1}}&{{w_2}}&{{w_3}}&{{w_4}}&0&0&0
\end{array}} \right]}_{\bf{B}}\underbrace {\left[ {\begin{array}{*{20}{c}}
{{\beta ^6}}\\
{{\beta ^5}}\\
{{\beta ^4}}\\
{{\beta ^3}}\\
{{\beta ^2}}\\
\beta \\
1
\end{array}} \right]}_{\bf{J}} = 0
	\end{aligned}.
 \label{38}
\end{equation}
\end{small}

In this case, Eq.~\eqref{38} can be rewritten as:
\begin{equation}
	\begin{aligned}
{{\bf{B}}_{7 \times 7}}{{\bf{J}}_{7 \times 1}} = {\bf{0}}
	\end{aligned},
	\label{39}
\end{equation}
where matrix B contains only unknown ${s}$. Eq.~\eqref{39} can be rewritten as:
\begin{equation}
	\begin{aligned}
({{\bf{B}}_0} + s{{\bf{B}}_1} + {s^2}{{\bf{B}}_2} +  \ldots  + {s^{16}}{{\bf{B}}_{16}}){\bf{J}} = 0
	\end{aligned}.
	\label{40}
\end{equation}

For the convenience of description, we define the matrices ${\bf{D}}$, ${\bf{Q}}$, and ${\bf{L}}$ as follows:
\begin{equation}
	\begin{aligned}
\begin{array}{l}
{\bf{D}} = \left[ {\begin{array}{*{20}{c}}
{\bf{0}}&{\bf{I}}& \ldots &{\bf{0}}\\
{\bf{0}}&{\bf{0}}& \ldots &{\bf{0}}\\
 \ldots & \ldots & \ldots &{\bf{I}}\\
{ - {{\bf{B}}_0}}&{ - {{\bf{B}}_1}}& \ldots &{ - {{\bf{B}}_{15}}}
\end{array}} \right],\\
{\bf{Q}} = \left[ {\begin{array}{*{20}{c}}
{\bf{I}}&{\bf{0}}& \ldots &{\bf{0}}\\
{\bf{0}}&{\bf{I}}& \ldots &{\bf{0}}\\
 \ldots & \ldots & \ldots &{\bf{0}}\\
{\bf{0}}&{\bf{0}}& \ldots &{{{\bf{B}}_{16}}}
\end{array}} \right],\\
{\bf{L}} = \left[ {\begin{array}{*{20}{c}}
{\bf{J}}\\
{s{\bf{J}}}\\
 \ldots \\
{{s^{15}}{\bf{J}}}
\end{array}} \right].
\end{array}
	\end{aligned}
	\label{41}
\end{equation}

From Eqs.~\eqref{40} and~\eqref{41}, we can get ${\bf{DL}} = ${s}${\bf{QL}}$. Hence, the eigenvalue of ${{\bf{Q}}^{ - 1}}{\bf{D}}$ is ${s}$, which can be written as follows:
\begin{small}
\begin{equation}
	\begin{aligned}
{{\bf{Q}}^{ - 1}}{\bf{D}} = \left[ {\begin{array}{*{20}{c}}
{\bf{0}}&{\bf{I}}& \ldots &{\bf{0}}\\
{\bf{0}}&{\bf{0}}& \ldots &{\bf{0}}\\
 \ldots & \ldots & \ldots &{\bf{I}}\\
{ - {\bf{B}}_{16}^{ - 1}{{\bf{B}}_0}}&{ - {\bf{B}}_{16}^{ - 1}{{\bf{B}}_1}}& \ldots &{ - {\bf{B}}_{16}^{ - 1}{{\bf{B}}_{15}}}
\end{array}} \right]
	\end{aligned}.
 \label{42}
\end{equation}
\end{small}

In this case, ${{\bf{B}}_{16}}$ is the singular matrix because the first column is full of zeros. ${{\bf{B}}_{0}}$ is full rank, and the inverse of ${{\bf{B}}_{0}}$ is more stable than ${{\bf{B}}_{16}}$. Hence, we define $z = \frac{1}{s}$. Eq.~\eqref{40} can be rewritten
\begin{small}
\begin{equation}
({{\mathop{\rm z}\nolimits} ^{16}}{{\bf{B}}_0} + {z^{15}}{{\bf{B}}_1} +  \ldots  + {{\bf{B}}_{16}}){\bf{J}} = 0.
\end{equation}
\label{43}
\end{small}
The eigenvalue of matrix ${\bf{G}}$ is ${z}$, which is expressed as
\begin{small}
\begin{equation}
	\begin{aligned}
{\bf{G}} = \left[ {\begin{array}{*{20}{c}}
{\bf{0}}&{\bf{I}}& \ldots &{\bf{0}}\\
{\bf{0}}&{\bf{0}}& \ldots &{\bf{0}}\\
 \ldots & \ldots & \ldots &{\bf{I}}\\
{ - {\bf{B}}_0^{ - 1}{{\bf{B}}_{16}}}&{ - {\bf{B}}_0^{ - 1}{{\bf{B}}_{15}}}& \ldots &{ - {\bf{B}}_0^{ - 1}{{\bf{B}}_1}}
\end{array}} \right]
	\end{aligned}.
 \label{44}
\end{equation}
\end{small}

Since the matrix ${\bf{G}}$ has null columns, this leads to the presence of zeros in the eigenvalues. These columns and corresponding rows are removed. The size of ${\bf{G}}$ is $88 \times 88$. We can obtain ${s}$ by using real Schur decomposition to solve the eigenvalue of ${{\bf{G}}}$. Finally, the rotation matrix can be obtained based on Eq.~\eqref{6}, and the translation vector can be obtained from the eigenvectors of ${\bf{C}}$. The original rotation matrix and translation vector can be obtained through Eq.~\eqref{11} and Eq.~\eqref{12}.
\section{Linearized Solver}
In some practical applications, the relative rotation between consecutive views of the multi-camera system is often small. When the yaw angle ${\theta _y}$ between two views is small, according to the limit theorem, we can obtain

\begin{equation}
\left\{ \begin{array}{l}
\mathop {{\rm{lim}}}\limits_{\theta {}_y \to 0} {\rm{sin}}(\theta {}_y ) = \theta {}_y, \\
\mathop {{\rm{lim}}}\limits_{\theta {}_y \to 0} {\rm{cos}}(\theta {}_y ) = 1.
\end{array} \right.
\end{equation}

Eq.~\eqref{6} can be rewritten as:
\begin{equation}
	\begin{aligned}
{{\bf{R}}_y} = \left[ {\begin{array}{*{20}{c}}
1&0&{{\theta _y}}\\
0&1&0\\
{ - {\theta _y}}&0&1
\end{array}} \right].
	\end{aligned}
\end{equation}

Similar to the above section, we can get that ${{g_1}}$, ${{g_2}}$, ${{g_3}}$, ${{g_4}}$, ${{h_1}}$, ${{h_2}}$, ${{h_3}}$, and ${{h_4}}$ are polynomials of ${s}$. Table~\ref{tab:xishu_} shows the highest degree of variables ${s}$. In this case, the size of ${\bf{G}}$ is $40 \times 40$ in Eq.~\eqref{44}. The method of estimating the rotation matrix and the translation vector can refer to the previous method in the above section.
\begin{table}
 \centering
  \caption{Degree of ${g_i},{w_i}$ when the angle is small (${i=1,2,3,4}$).}
 \small
\begin{tabular}{ccccccccc} 
\hline
\toprule     
    & ${g_1}$ & ${g_2}$  & ${g_3}$& ${g_4}$ & ${w_1}$ & ${w_2}$  & ${w_3}$ & ${w_4}$  \\
\midrule 
Degree(s) & 2 & 4 & 6 & 8 & 1 & 3 & 5 & 7\\
\bottomrule  
\end{tabular}
 \label{tab:xishu_}
\end{table}

\section{Experiments}
We test the accuracy of the proposed solver on simulated data and real data, respectively. The solver is compared with state-of-the-art methods, including \verb+4pt-Lee+~\cite{4pt_lee}, \verb+4pt-Liu+~\cite{4pt_liu}, \verb+4pt-Sweeney+~\cite{4pt_Sweeney}, and \verb+Wu+~\cite{Ding_mul}. Our method is named \verb+OURS+. All methods use the known vertical direction as a prior. The angular difference between the estimated rotation and the truth rotation as rotation error ${\varepsilon _{\bf{R}}}$. We use ${\varepsilon _{\bf{t}}}$ and ${\varepsilon _{{\bf{t}},{\rm{dir}}}}$ to evaluate the accuracy of the translation vector. ${\varepsilon _{\bf{t}}}$ is a scalar about the translation error~\cite{quan1999linear} and ${\varepsilon _{{\bf{t}},{\rm{dir}}}}$ represents the direction of translation. ${\varepsilon _{\bf{R}}}$, ${\varepsilon _{\bf{t}}}$, and ${\varepsilon _{{\bf{t}},{\rm{dir}}}}$ can be written as:

$ \bullet $ ${\varepsilon _{\bf{R}}} = {\rm{arccos}}(\frac{{{\rm{trace}}({{\bf{R}}_{gt}}{{\bf{R}}^T}) - 1}}{2})$,

$ \bullet $ ${\varepsilon _{\bf{t}}} = 2\frac{{\left\| {{{\bf{t}}_{gt}} - {\bf{t}}} \right\|}}{{(\left\| {{{\bf{t}}_{gt}}} \right\| + \left\| {\bf{t}} \right\|)}}$,

$ \bullet $ ${\varepsilon _{{\bf{t}},{\rm{dir}}}} = {\rm{arccos}}\frac{{{\bf{t}}_{gt}^T{\bf{t}}}}{{(\left\| {{{\bf{t}}_{gt}}} \right\| + \left\| {\bf{t}} \right\|)}}$,\\
where ${{{\bf{R}}_{gt}}}$ and ${{{{\bf{t}}_{gt}}}}$ are the ground truth of rotation and translation, respectively. ${{{\bf{R}}}}$ and ${{{{\bf{t}}}}}$ are the estimated rotation and estimated translation, respectively.

\begin{figure*}[htbp]       
  \centering
     \subfloat[${\varepsilon _{\bf{R}}}$]{
     \begin{minipage}[t]{0.24\linewidth}
     \centering
     \includegraphics[width=0.9\linewidth]{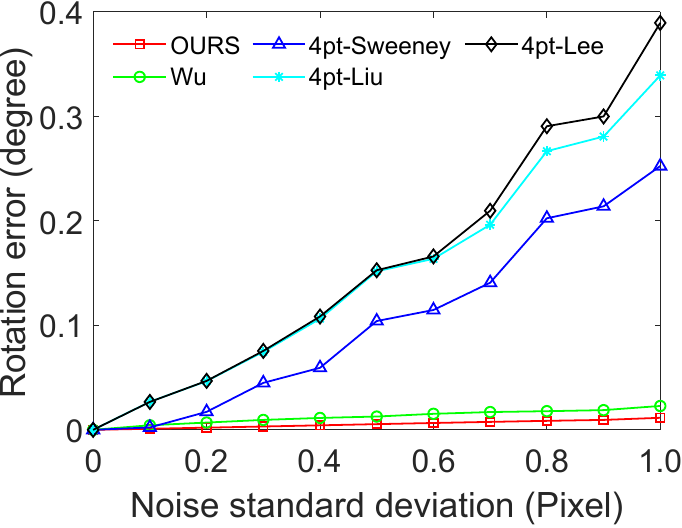}
     \label{fig.3a}
     \end{minipage}
      }
     \subfloat[${\varepsilon _{\bf{R}}}$]{
     \begin{minipage}[t]{0.24\linewidth}
     \centering
     \includegraphics[width=0.9\linewidth]{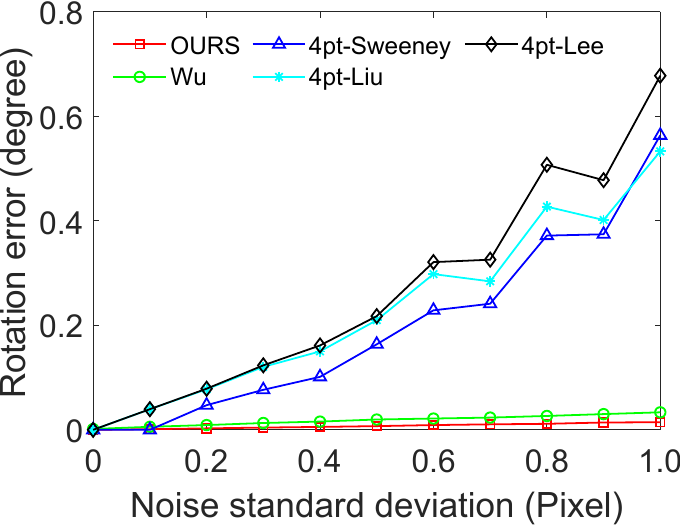}
    \label{fig.3b}
     \end{minipage}
      }
     \subfloat[${\varepsilon _{\bf{R}}}$]{
     \begin{minipage}[t]{0.24\linewidth}
     \centering
     \includegraphics[width=0.9\linewidth]{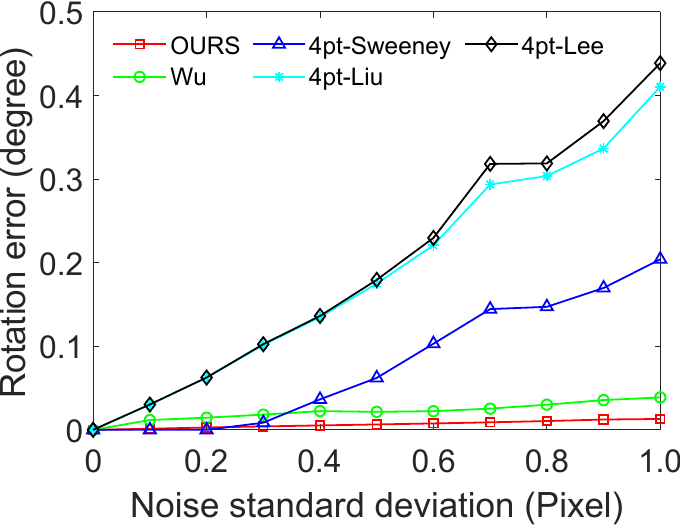}
     \label{fig.3c}
     \end{minipage}
      }
     \subfloat[${\varepsilon _{\bf{R}}}$]{
     \begin{minipage}[t]{0.24\linewidth}
     \centering
     \includegraphics[width=0.9\linewidth]{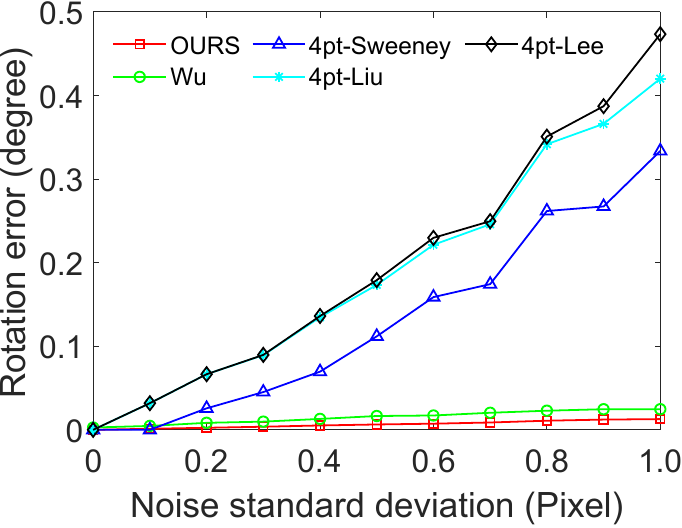}
     \label{fig.3c_1}
     \end{minipage}
      }
   \vspace{-4pt}
      
    \subfloat[${\varepsilon _{{\bf{t}},{\rm{dir}}}}$]{  
     \begin{minipage}[t]{0.24\linewidth}
     \centering
     \includegraphics[width=0.9\linewidth]{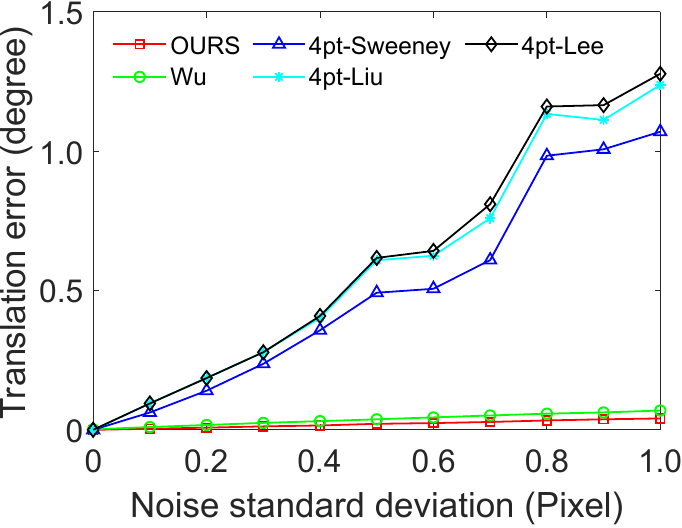}
     \label{fig.3d}
     \end{minipage}
      }
    \subfloat[${\varepsilon _{{\bf{t}},{\rm{dir}}}}$]{  
     \begin{minipage}[t]{0.24\linewidth}
     \centering
     \includegraphics[width=0.9\linewidth]{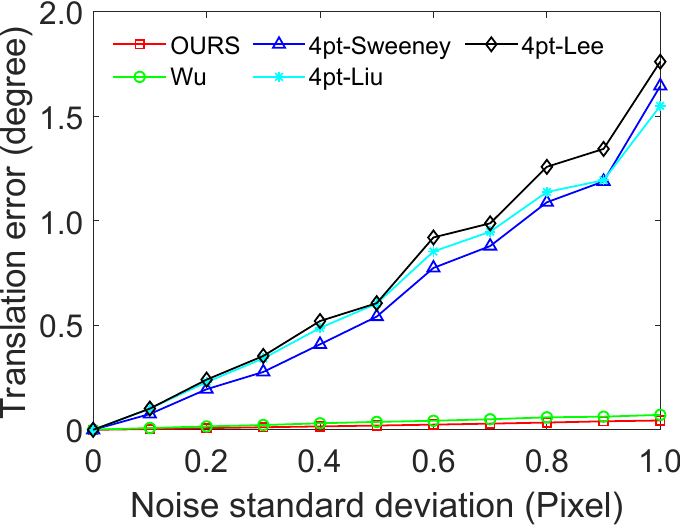}
    \label{fig.3e}
     \end{minipage}
      } 
    \subfloat[${\varepsilon _{{\bf{t}},{\rm{dir}}}}$]{  
     \begin{minipage}[t]{0.24\linewidth}
     \centering
     \includegraphics[width=0.9\linewidth]{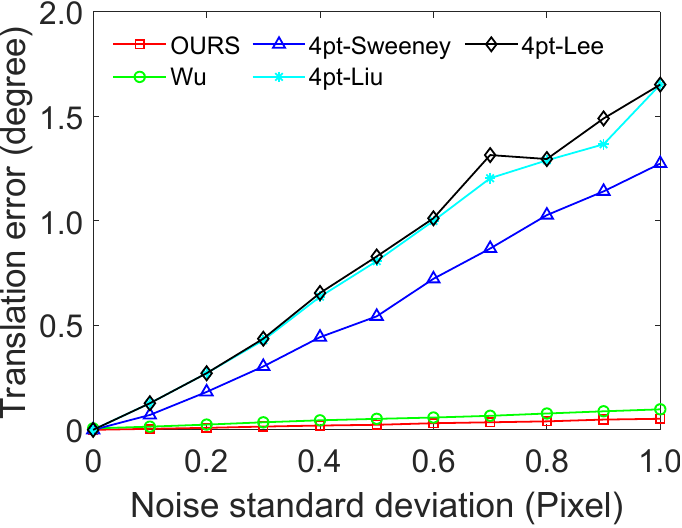}
     \label{fig.3f}
     \end{minipage}
      } 
    \subfloat[${\varepsilon _{{\bf{t}},{\rm{dir}}}}$]{  
     \begin{minipage}[t]{0.24\linewidth}
     \centering
     \includegraphics[width=0.9\linewidth]{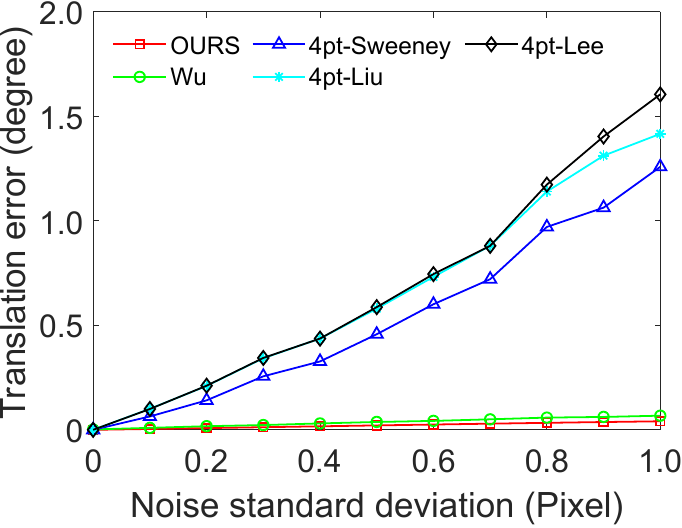}
     \label{fig.3g}
     \end{minipage}
      }  
      \vspace{-4pt}

     \subfloat[${\varepsilon _{\bf{t}}}$]{
     \begin{minipage}[t]{0.24\linewidth}
     \centering
     \includegraphics[width=0.9\linewidth]{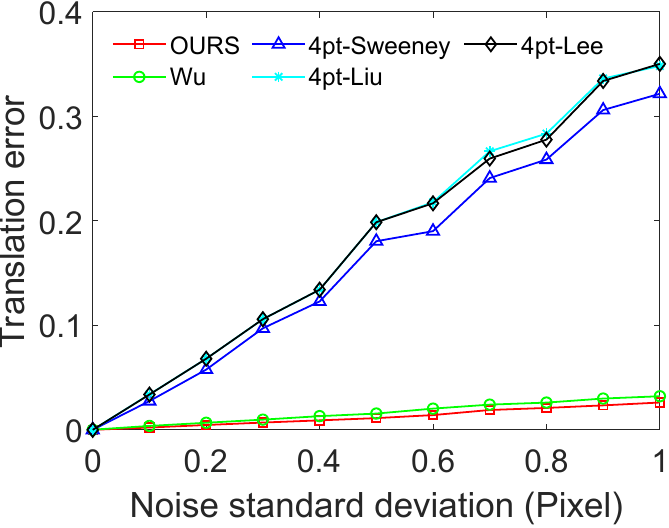}
     \label{fig.3h}
     \end{minipage} 
      } 
   \subfloat[${\varepsilon _{\bf{t}}}$]{
     \begin{minipage}[t]{0.24\linewidth}
     \centering
     \includegraphics[width=0.9\linewidth]{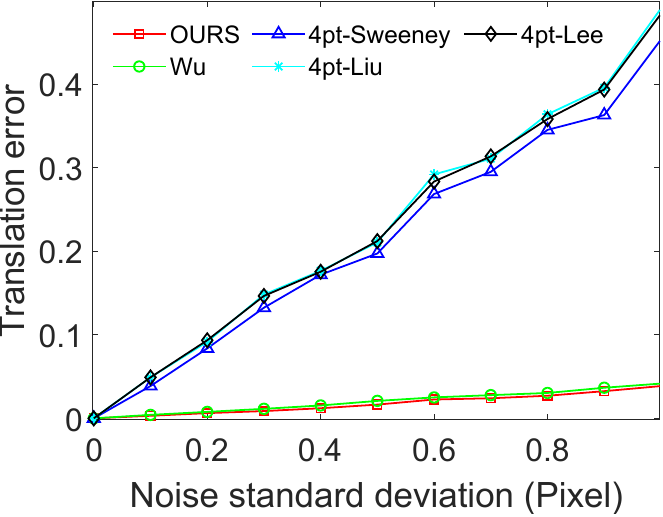}
    \label{fig.3i}
     \end{minipage} 
      }  
     \subfloat[${\varepsilon _{\bf{t}}}$]{
     \begin{minipage}[t]{0.24\linewidth}
     \centering
     \includegraphics[width=0.9\linewidth]{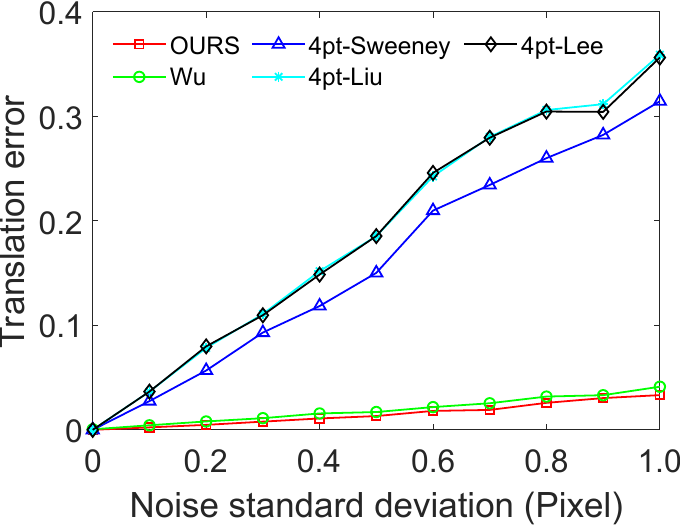}
     \label{fig.3j}
     \end{minipage}
      } 
    \subfloat[${\varepsilon _{\bf{t}}}$]{
     \begin{minipage}[t]{0.24\linewidth}
     \centering
     \includegraphics[width=0.9\linewidth]{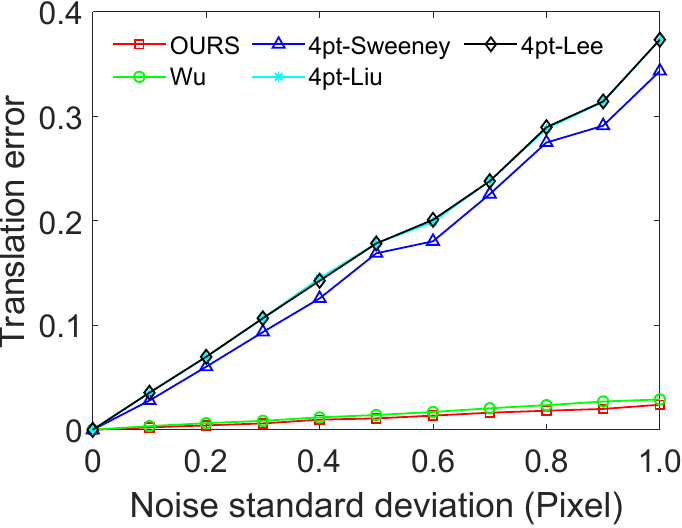}
     \label{fig.3k}
     \end{minipage}
      }  
    \centering
    \caption{Add noise to image pixels (unit: pixel) in four motion modes. The first column: random motion; The second column: forward motion; The third column: planar motion; The fourth column: sideways motion. }
     \vspace{-10pt}
    \label{fig.3}
  
\end{figure*}

\begin{figure*}[ttp]            
  \centering
     \subfloat[${\varepsilon _{\bf{R}}}$]{
     \begin{minipage}[t]{0.24\linewidth}
     \centering
     \includegraphics[width=0.9\linewidth]{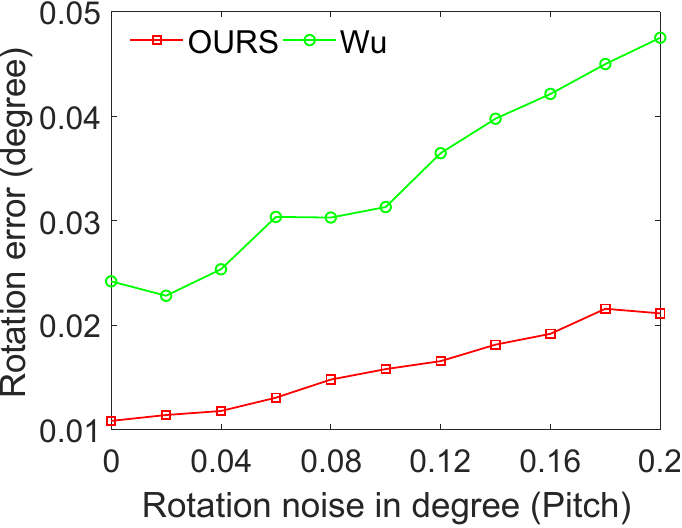}
     \label{fig.4a}
     \end{minipage}
      }
     \subfloat[${\varepsilon _{\bf{R}}}$]{
     \begin{minipage}[t]{0.24\linewidth}
     \centering
     \includegraphics[width=0.9\linewidth]{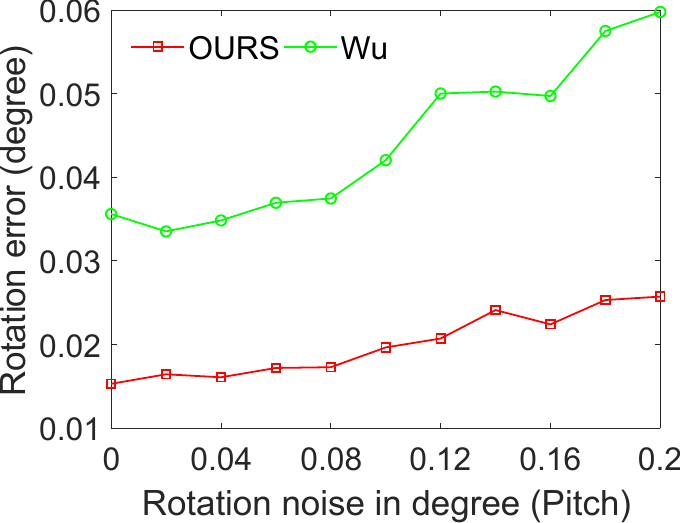}
    \label{fig.4b}
     \end{minipage}
      } 
      \subfloat[${\varepsilon _{\bf{R}}}$]{
     \begin{minipage}[t]{0.24\linewidth}
     \centering
     \includegraphics[width=0.9\linewidth]{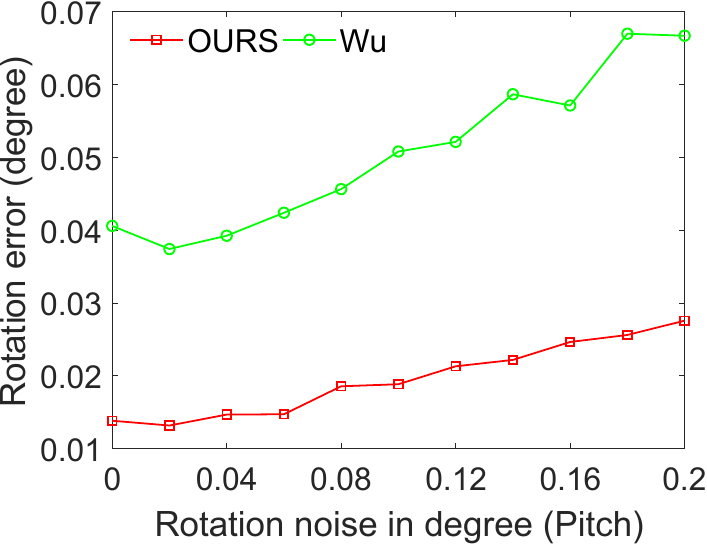}
     \label{fig.4c}
     \end{minipage}
      } 
    \subfloat[${\varepsilon _{\bf{R}}}$]{
     \begin{minipage}[t]{0.24\linewidth}
     \centering
     \includegraphics[width=0.9\linewidth]{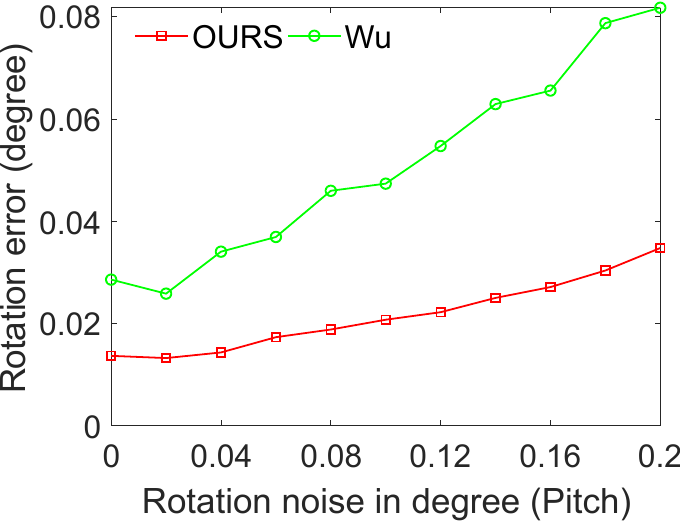}
     \label{fig.4d}
     \end{minipage}
      } 
    \vspace{-4pt}
      
    \subfloat[${\varepsilon _{{\bf{t}},{\rm{dir}}}}$]{
     \begin{minipage}[t]{0.24\linewidth}
     \centering
     \includegraphics[width=0.9\linewidth]{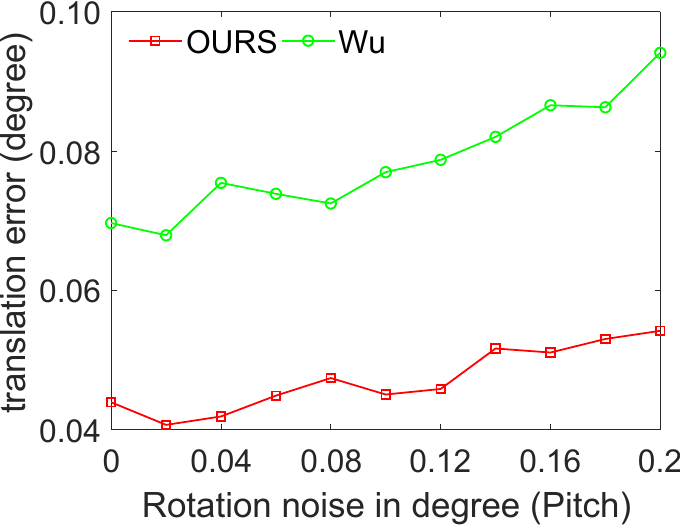}
     \label{fig.4e}
     \end{minipage}
      }
    \subfloat[${\varepsilon _{{\bf{t}},{\rm{dir}}}}$]{
     \begin{minipage}[t]{0.24\linewidth}
     \centering
     \includegraphics[width=0.9\linewidth]{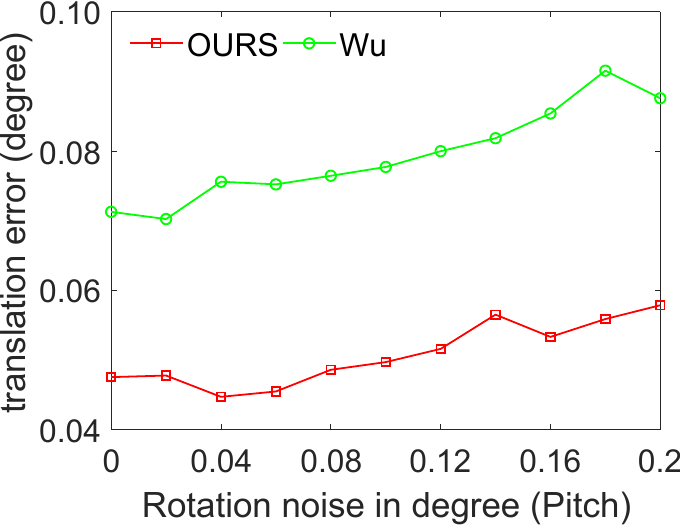}
    \label{fig.4f}
     \end{minipage}
      } 
     \subfloat[${\varepsilon _{{\bf{t}},{\rm{dir}}}}$]{
     \begin{minipage}[t]{0.24\linewidth}
     \centering
     \includegraphics[width=0.9\linewidth]{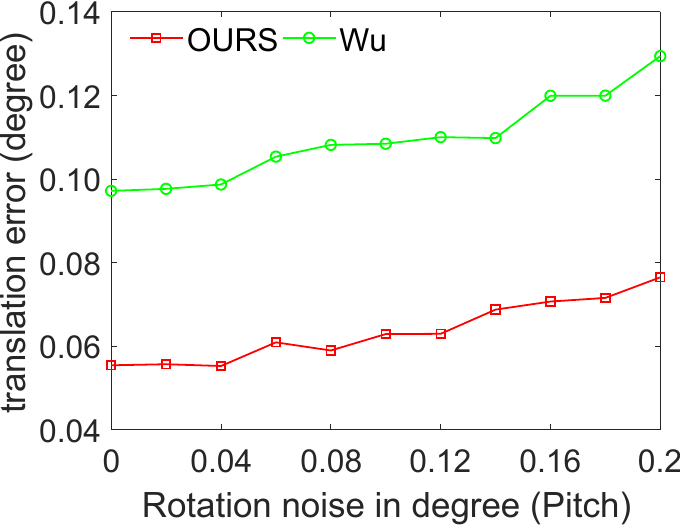}
     \label{fig.4g}
     \end{minipage}
      } 
     \subfloat[${\varepsilon _{{\bf{t}},{\rm{dir}}}}$]{
     \begin{minipage}[t]{0.24\linewidth}
     \centering
     \includegraphics[width=0.9\linewidth]{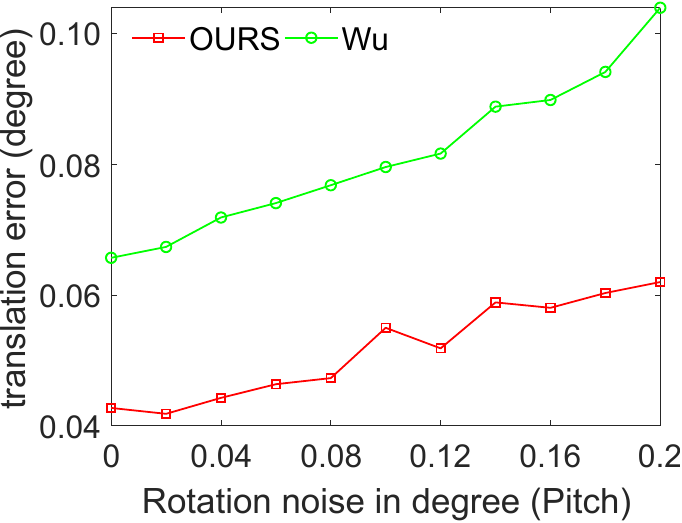}
     \label{fig.4h}
     \end{minipage}
      } 
   \vspace{-4pt}

     \subfloat[${\varepsilon _{\bf{t}}}$]{
     \begin{minipage}[t]{0.24\linewidth}
     \centering
     \includegraphics[width=0.9\linewidth]{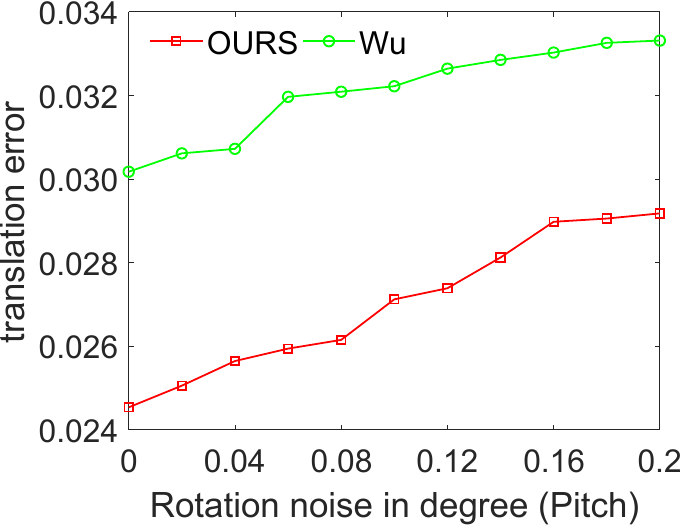}
     \label{fig.4i}
     \end{minipage}
      }
    \subfloat[${\varepsilon _{\bf{t}}}$]{
     \begin{minipage}[t]{0.24\linewidth}
     \centering
     \includegraphics[width=0.9\linewidth]{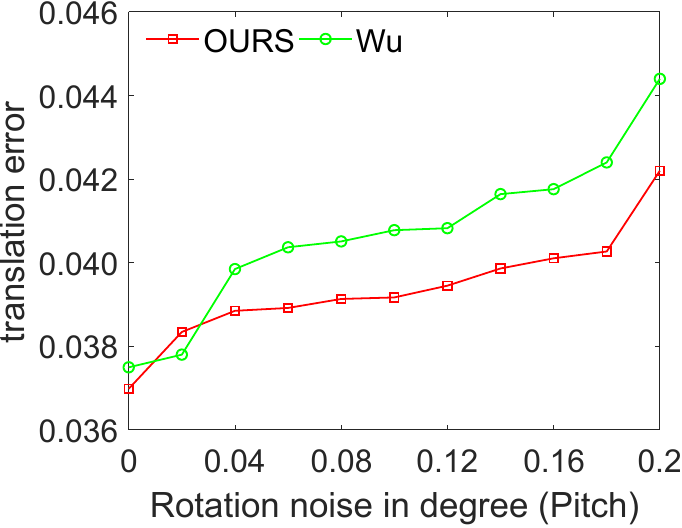}
    \label{fig.4j}
     \end{minipage}
      } 
     \subfloat[${\varepsilon _{\bf{t}}}$]{
     \begin{minipage}[t]{0.24\linewidth}
     \centering
     \includegraphics[width=0.9\linewidth]{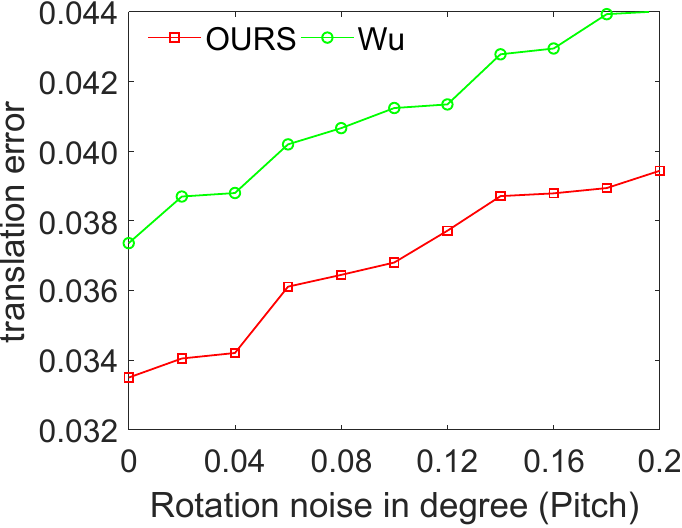}
     \label{fig.4k}
     \end{minipage}
      } 
     \subfloat[${\varepsilon _{\bf{t}}}$]{
     \begin{minipage}[t]{0.24\linewidth}
     \centering
     \includegraphics[width=0.9\linewidth]{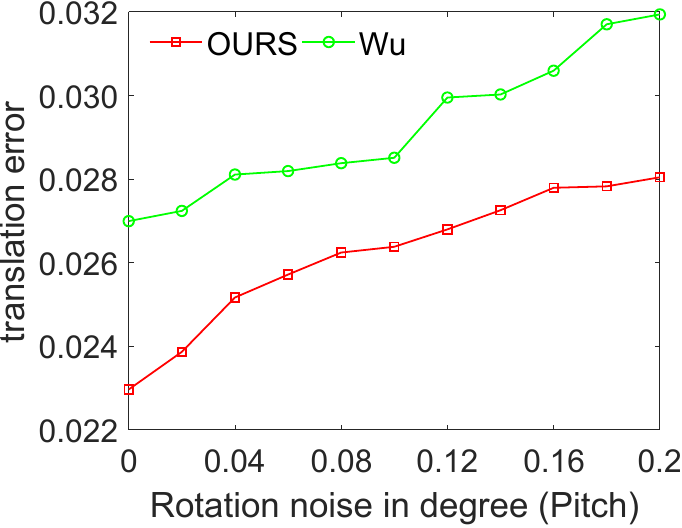}
     \label{fig.4l}
     \end{minipage}
      } 
   \vspace{-4pt}
    \centering
    \caption{ Add noise to pitch degree (unit: degree) in four motion modes. The first column: random motion; The second column: forward motion; The third column: planar motion; The fourth column: sideways motion.}
    \label{fig.4}
   \vspace{-10pt}
\end{figure*}

\begin{figure*}[htbp]            
  \centering
     \subfloat[${\varepsilon _{\bf{R}}}$]{
     \begin{minipage}[t]{0.24\linewidth}
     \centering
     \includegraphics[width=0.9\linewidth]{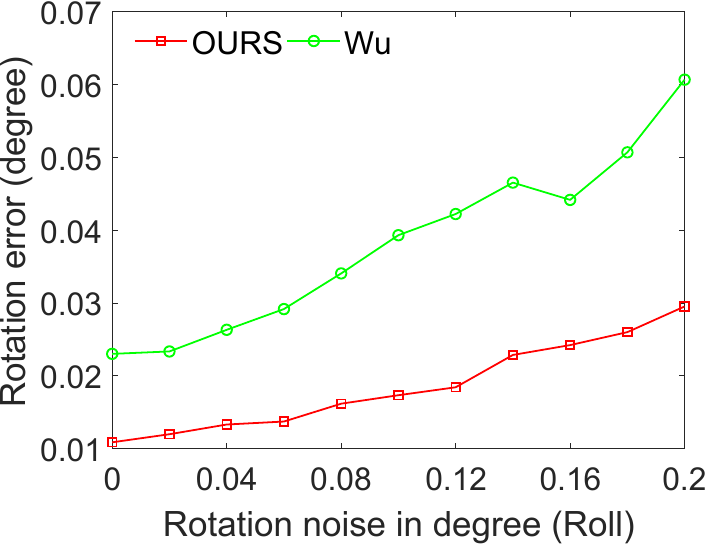}
     \label{fig.5a}
     \end{minipage}
      }
    \subfloat[${\varepsilon _{\bf{R}}}$]{
     \begin{minipage}[t]{0.24\linewidth}
     \centering
     \includegraphics[width=0.9\linewidth]{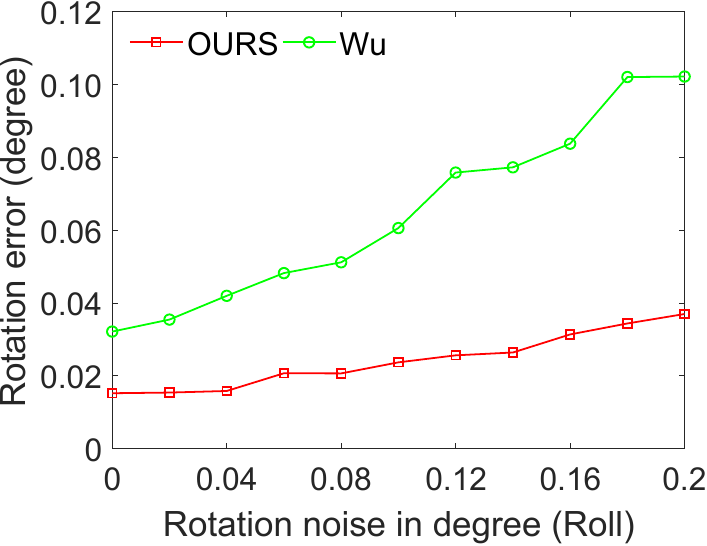}
    \label{fig.5b}
     \end{minipage}
      } 
     \subfloat[${\varepsilon _{\bf{R}}}$]{
     \begin{minipage}[t]{0.24\linewidth}
     \centering
     \includegraphics[width=0.9\linewidth]{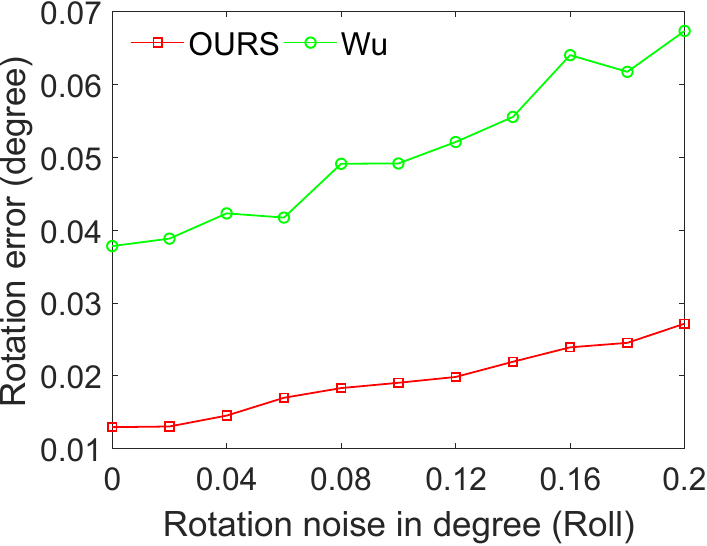}
     \label{fig.5c}
     \end{minipage}
      } 
      \subfloat[${\varepsilon _{\bf{R}}}$]{
     \begin{minipage}[t]{0.24\linewidth}
     \centering
     \includegraphics[width=0.9\linewidth]{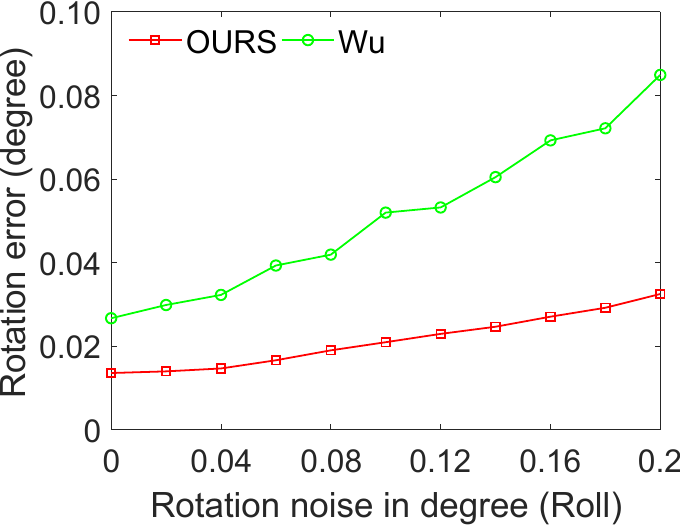}
     \label{fig.5d}
     \end{minipage}
      } 
   \vspace{-4pt}

     \subfloat[${\varepsilon _{{\bf{t}},{\rm{dir}}}}$]{
     \begin{minipage}[t]{0.24\linewidth}
     \centering
     \includegraphics[width=0.9\linewidth]{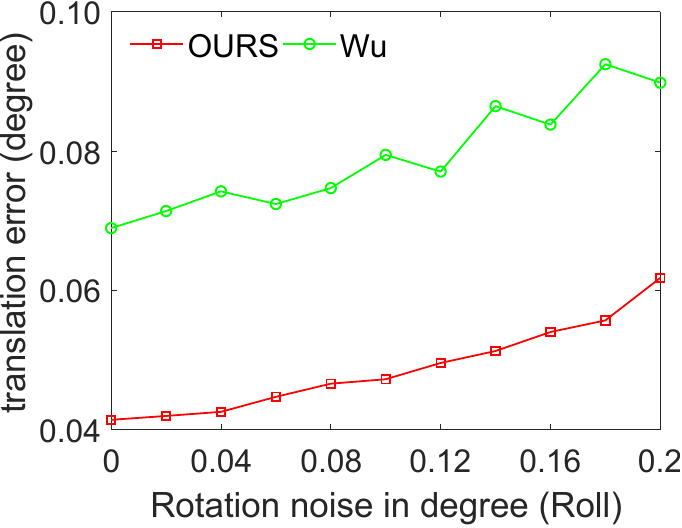}
     \label{fig.5e}
     \end{minipage}
      }
    \subfloat[${\varepsilon _{{\bf{t}},{\rm{dir}}}}$]{
     \begin{minipage}[t]{0.24\linewidth}
     \centering
     \includegraphics[width=0.9\linewidth]{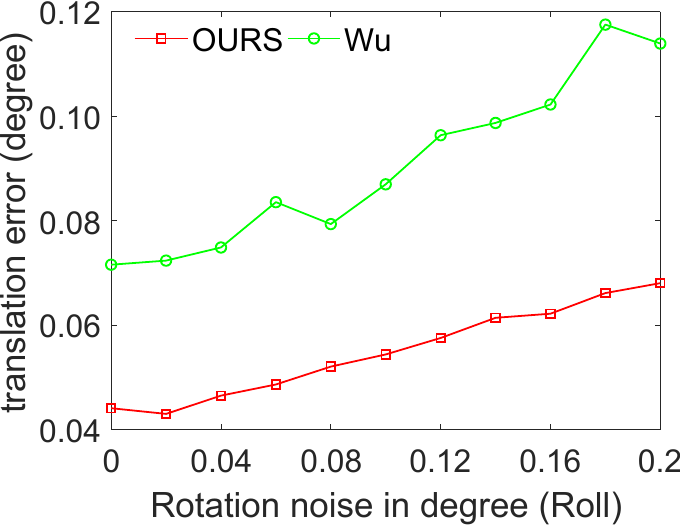}
    \label{fig.5f}
     \end{minipage}
      } 
      \subfloat[${\varepsilon _{{\bf{t}},{\rm{dir}}}}$]{
     \begin{minipage}[t]{0.24\linewidth}
     \centering
     \includegraphics[width=0.9\linewidth]{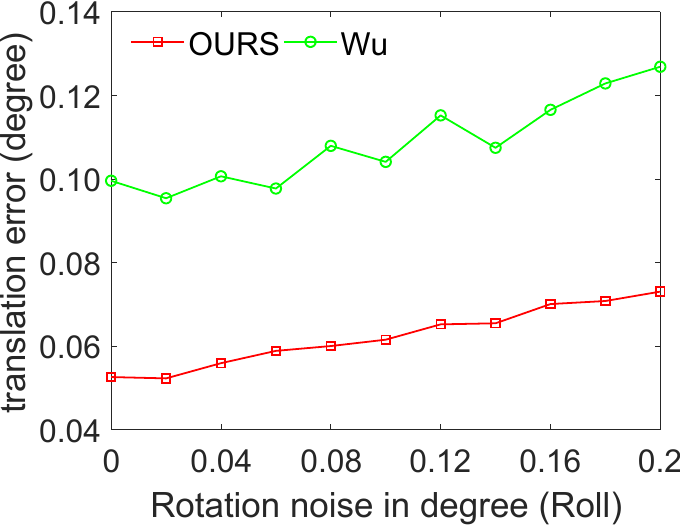}
     \label{fig.5g}
     \end{minipage}
      } 
      \subfloat[${\varepsilon _{{\bf{t}},{\rm{dir}}}}$]{
     \begin{minipage}[t]{0.24\linewidth}
     \centering
     \includegraphics[width=0.9\linewidth]{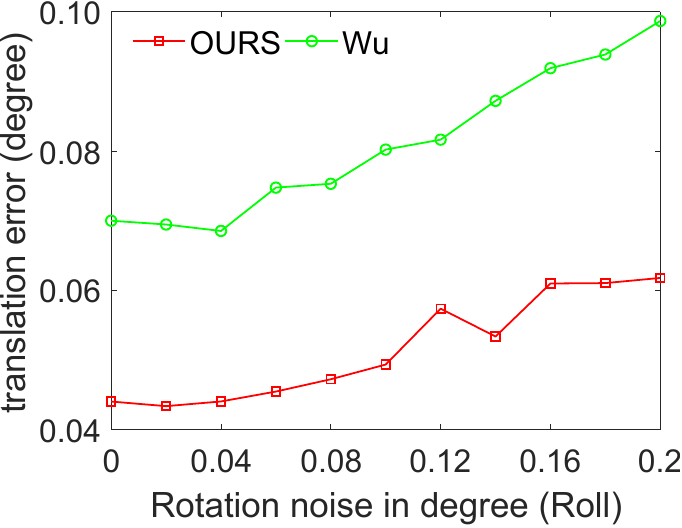}
     \label{fig.5h}
     \end{minipage}
      } 
   \vspace{-4pt}
   
      \subfloat[${\varepsilon _{\bf{t}}}$]{
     \begin{minipage}[t]{0.24\linewidth}
     \centering
     \includegraphics[width=0.9\linewidth]{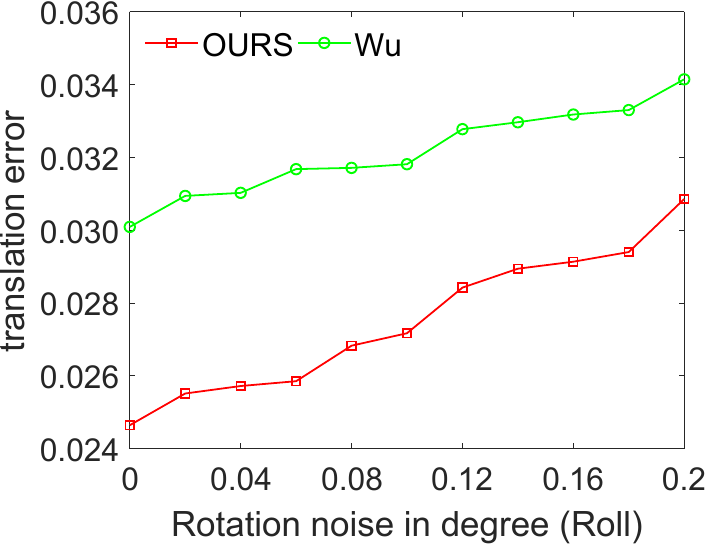}
     \label{fig.5i}
     \end{minipage}
      }
      \subfloat[${\varepsilon _{\bf{t}}}$]{
     \begin{minipage}[t]{0.24\linewidth}
     \centering
     \includegraphics[width=0.9\linewidth]{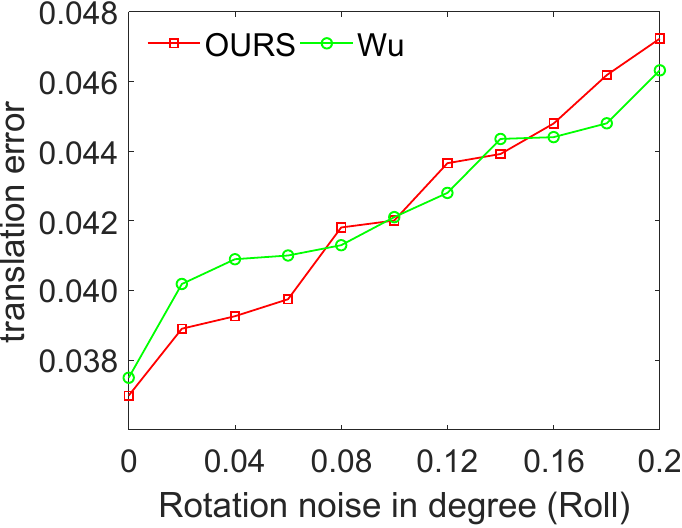}
    \label{fig.5j}
     \end{minipage}
      } 
     \subfloat[${\varepsilon _{\bf{t}}}$]{
     \begin{minipage}[t]{0.24\linewidth}
     \centering
     \includegraphics[width=0.9\linewidth]{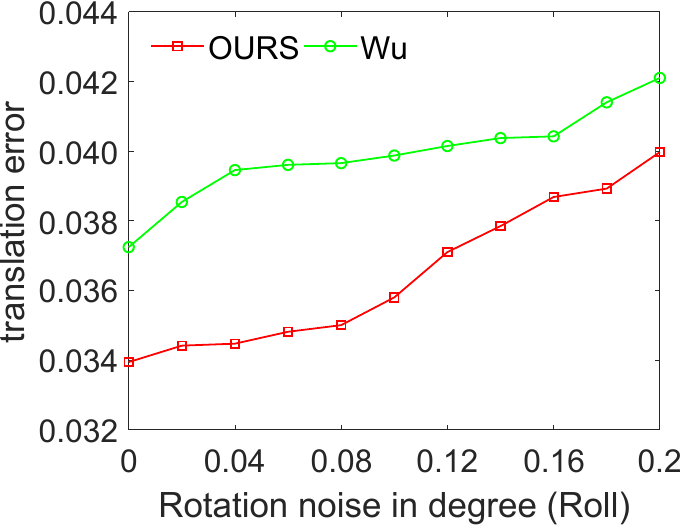}
     \label{fig.5k}
     \end{minipage}
      } 
      \subfloat[${\varepsilon _{\bf{t}}}$]{
     \begin{minipage}[t]{0.24\linewidth}
     \centering
     \includegraphics[width=0.9\linewidth]{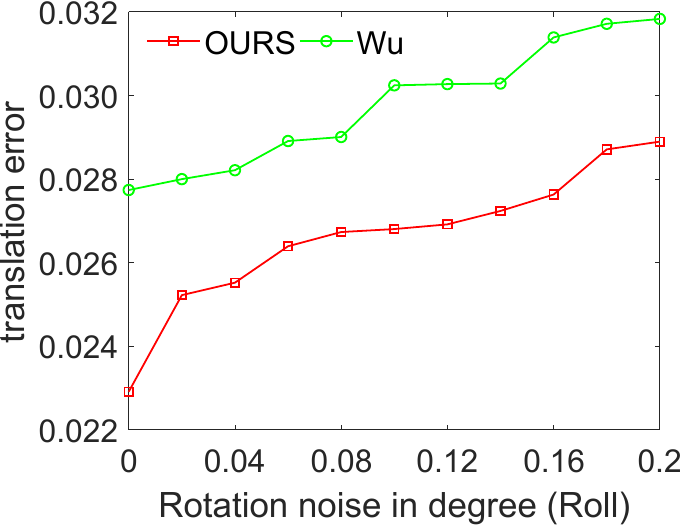}
     \label{fig.5l}
     \end{minipage}
      } 
   \vspace{-4pt}

    \centering
    \caption{ Add noise to roll degree (unit: degree) in four motion modes. The first column: random motion; The second column: forward motion; The third column: planar motion; The fourth column: sideways motion.}
    \label{fig.5}
   \vspace{-10pt}
\end{figure*}

\subsection{Synthetic Evaluation}
We simulate a multi-camera system consisting of four cameras~\cite{kneip2014opengv}. The multi-camera reference system is built in the middle of the camera rig. The distance from each camera to the middle of the camera rig is 0.5m. We randomly generate 100 planes around the multi-camera system, then take a 3D point on each plane and project the 3D point onto four cameras. The resolution of the cameras is $640 \times 480$ pixels. The principal points are (320, 240) pixels. The focal lengths of cameras are 400 pixels. In the synthetic experiment, 1000 trials are carried out. 

The method proposed in this paper, along with the comparison solvers, is tested with added image noise and IMU noise under four different motions: random (${\bf{t}} = {\left[ {\begin{array}{*{20}{c}}{{t_x}}&{{t_y}}&{{t_z}}\end{array}} \right]^T}$), planar (${\bf{t}} = {\left[ {\begin{array}{*{20}{c}}{{t_x}}&{{0}}&{{t_z}}\end{array}} \right]^T}$), forward (${\bf{t}} = {\left[ {\begin{array}{*{20}{c}}{{0}}&{{0}}&{{t_z}}\end{array}} \right]^T}$), and sideways (${\bf{t}} = {\left[ {\begin{array}{*{20}{c}}{{t_x}}&{{0}}&{{0}}\end{array}} \right]^T}$). The angle of rotation angle is randomly chosen from ${-10^{\rm{^\circ }}}$ to ${10^{\rm{^\circ }}}$. The image noise is set to 1 pixel when adding noise to the IMU. Since the noise value of accelerometers in cars and smartphones is around ${0.2^{\rm{^\circ }}}$, the maximum value of noise on pitch and roll is ${0.2^{\rm{^\circ }}}$~\cite{ding2021globally}. We add noise to the local affine transformation using the method described in~\cite{barath2019homography}. Local affine transformations can be obtained by a first-order approximation of the homography matrix.
\begin{equation}
	\begin{aligned}
          	{{a_{11}} = \frac{{{h_{11}} - {h_{31}}{u_j }}}{b}},{\qquad \quad}&{{a_{12}} = \frac{{{h_{21}} - {h_{31}}{v_j }}}{b}},\\
     	{{a_{21}} = \frac{{{h_{12}} - {h_{32}}{u_j }}}{b}},{\qquad \quad}&{{a_{22}} =    \frac{{{h_{22}} - {h_{32}}{v_j }}}{b}},
	\end{aligned}
\end{equation}
where $b = {\bf{h}}_3^T{\left[ {\begin{array}{*{20}{l}} {{u_i}}&{{v_i}}&1 \end{array}} \right]^T}$ , and ${\bf{h}}_3^T$ represents the last row of the homography. ${h_{11}}$, ${h_{12}}$, ${h_{21}}$, ${h_{22}}$, ${h_{31}}$, and ${h_{32}}$ represent the elements of the homography matrix.

$\bf{Pixel}$ $\bf{Noise}$ $\bf{Resilience}$: Fig.~\ref{fig.3} shows the performance of the \verb+4pt-Lee+, \verb+4pt-Liu+, \verb+4pt-Sweeney+, \verb+Wu+, and \verb+OURS+ methods when image noise is added. The first, second, third, and fourth columns of Fig.~\ref{fig.3} represent the experimental result for random motion, forward motion, plane motion, and sideways motion, respectively. The  performance of the \verb+OURS+ method is better than \verb+4pt-Lee+, \verb+4pt-Liu+, \verb+4pt-Sweeney+, and \verb+Wu+ in the four motion models. The performance of the \verb+Wu+ method is better than that of \verb+4pt-Lee+, \verb+4pt-Liu+, and \verb+4pt-Sweeney+ in estimating the translation vector in Fig.~\ref{fig.3}(e)-(l).

$\bf{IMU}$ $\bf{Noise}$ $\bf{Resilience}$: From Fig.~\ref{fig.3}, it is evident that the \verb+Wu+ and \verb+OURS+ methods show significantly better performance in calculating translation and rotation compared to the \verb+4pt-Lee+, \verb+4pt-Liu+, and \verb+4pt-Sweeney+ methods when the image noise value is 1.0 pixel. To highlight the performance of our method, we only use the \verb+Wu+ method as a comparison method when IMU noise is added. Noise is added separately to the pitch and roll angles.

Fig.~\ref{fig.4} shows the performance of the \verb+Wu+ and \verb+OURS+ methods when pitch angle noise is added. Fig.~\ref{fig.5} shows the performance with added roll angle noise. The first, second, third, and fourth columns of Fig.~\ref{fig.4} and Fig.~\ref{fig.5} correspond to experimental results for random motion, forward motion, plane motion, and sideways motion, respectively. Our observations are as follows: (1)The \verb+OURS+ method exhibit notably better rotation calculation performance than the \verb+Wu+ method across all four motion modes when either pitch angle noise (Fig.~\ref{fig.4}(a)-(d)) or roll angle noise (Fig.~\ref{fig.5}(a)-(d)) is added. (2) The value of the ${\varepsilon _{{\bf{t}},{\rm{dir}}}}$ estimated by the \verb+OURS+ method is lower than that estimated by the \verb+Wu+ method across all motion modes when either pitch angle noise (Fig.~\ref{fig.4}(e)-(h)) or roll angle noise (Fig.~\ref{fig.5}(e)-(h)) is added. (3) The value of the ${\varepsilon _{{\bf{t}}}}$ estimated by the \verb+OURS+ method is less than that estimated by the \verb+Wu+ method in four modes of motion when pitch angle noise is added, except for forward motion at ${0.02^{\rm{^\circ }}}$ in Fig.~\ref{fig.4}(i)-(l). (4) The value of the ${\varepsilon _{{\bf{t}}}}$ estimated by the \verb+OURS+ method is lower than that estimated by the \verb+Wu+ method in four modes of motion when pitch angle noise is added, except for forward motion in Fig.~\ref{fig.5}(i)-(l). The \verb+OURS+ method shows improved translation calculation performance compared to the \verb+Wu+ method when roll angle noise is less than ${0.08^{\rm{^\circ }}}$ in Fig.~\ref{fig.5}(j).

\begin{table*}[tbp]
 \scriptsize{
  	\caption{Ablation experiments on random motion (degree)}
	\setlength{\tabcolsep}{1.5mm}{
			\scalebox{1.6}{
				\begin{tabular}{c c c c c c}
					\hline
					\multirow{2}{*}{\scriptsize{Seq.}}   
					& \scriptsize{\scriptsize{4pt-Lee}\cite{4pt_lee}}  
                       & \  \scriptsize{\scriptsize{4pt-Liu}\cite{4pt_liu}}  
                       & \ \scriptsize{\scriptsize{4pt-Sweeney}\cite{4pt_Sweeney}}
                       & \ \scriptsize{\scriptsize{Wu}\cite{Ding_mul}}  
                        & \scriptsize{\textbf{OURS}} \\
					\cline{2-6}
				   & \ \ ${\varepsilon _{\tiny{{\bf{R}}}}}$\quad ${\varepsilon _{{\bf{t}},{\rm{dir}}}}$      
                    & \ \ ${\varepsilon _{\tiny{{\bf{R}}}}}$ \quad ${\varepsilon _{{\bf{t}},{\rm{dir}}}}$      
                    & \ \ ${\varepsilon _{\tiny{{\bf{R}}}}}$ \quad ${\varepsilon _{{\bf{t}},{\rm{dir}}}}$      
                    & \ \ ${\varepsilon _{\tiny{{\bf{R}}}}}$ \quad ${\varepsilon _{{\bf{t}},{\rm{dir}}}}$    
                    & \ \ ${\varepsilon _{\tiny{{\bf{R}}}}}$ \quad ${\varepsilon _{{\bf{t}},{\rm{dir}}}}$  \\
					\hline
					Point& 0.3891  \  1.2705& 0.3491   \  1.2540&0.2510  \ 1.0545& 0.0277   \ 0.0698&   \textbf{0.0144} \  \textbf{0.0418} \\
					Pitch& 0.5615    \  1.1483& 0.4546   \  1.0235&	0.2374  \ 0.3575& 0.0352   \ 0.0827&	  \textbf{0.0141}  \  \textbf{0.0466} \\
					Roll& 0.4712   \  1.4228&	0.3872   \  1.1446&	0.2451  \ 0.4068& 0.0495  \ 0.0849&   \textbf{0.0201}  \  \textbf{0.0570} \\
					Rotation& 0.2015    \  0.9924& 0.3311   \  1.3301&	0.4823  \ 1.7687& 0.0423   \ 0.2670&	 \textbf{0.0298}       \  \textbf{0.2090} \\
                    Translation& 0.0065    \  0.0487& 0.0284   \  0.0687&	0.0109  \ 0.0586& 0.0005  \ 0.0102&	  \textbf{0.0004}  \  \textbf{0.0071} \\
					\hline  
		    \end{tabular}}
      }
      \label{tab:random}
 }
\end{table*}

 \begin{table*}[tbp]
 \scriptsize{
 	\caption{Ablation experiments on forward motion (degree)}
	\setlength{\tabcolsep}{1.5mm}{
			\scalebox{1.6}{
				\begin{tabular}{c c c c c c}
					\hline
					\multirow{2}{*}{\scriptsize{Seq.}}   
					& \scriptsize{\scriptsize{4pt-Lee}\cite{4pt_lee}}  
                       & \  \scriptsize{\scriptsize{4pt-Liu}\cite{4pt_liu}}  
                       & \ \scriptsize{\scriptsize{4pt-Sweeney}\cite{4pt_Sweeney}}
                       & \ \scriptsize{\scriptsize{Wu}\cite{Ding_mul}}  
                        & \scriptsize{\textbf{OURS}} \\
					\cline{2-6}
				   & \ \ ${\varepsilon _{\tiny{{\bf{R}}}}}$\quad ${\varepsilon _{{\bf{t}},{\rm{dir}}}}$      
                    & \ \ ${\varepsilon _{\tiny{{\bf{R}}}}}$ \quad ${\varepsilon _{{\bf{t}},{\rm{dir}}}}$      
                    & \ \ ${\varepsilon _{\tiny{{\bf{R}}}}}$ \quad ${\varepsilon _{{\bf{t}},{\rm{dir}}}}$      
                    & \ \ ${\varepsilon _{\tiny{{\bf{R}}}}}$ \quad ${\varepsilon _{{\bf{t}},{\rm{dir}}}}$    
                    & \ \ ${\varepsilon _{\tiny{{\bf{R}}}}}$ \quad ${\varepsilon _{{\bf{t}},{\rm{dir}}}}$  \\
					\hline
					Point& 0.6482  \  1.7211& 0.5791   \  1.6121&0.5850  \ 1.6851& 0.0332   \ 0.0720&   \textbf{0.0182} \  \textbf{0.0527} \\
					Pitch& 0.3948    \  0.6812& 0.3778   \  0.6333&	0.2523  \ 0.3152& 0.0447   \ 0.0798&	  \textbf{0.0227}  \  \textbf{0.0487} \\
					Roll& 0.3928    \  0.6352&	0.8011   \  1.0166&	0.3774  \ 0.5749& 0.0847  \ 0.1086&   \textbf{0.0319}  \  \textbf{0.0593} \\
					Rotation& 0.2682    \  1.0480& 0.2201   \ 0.8398&	0.3462  \ 1.0819& 0.0662   \ 0.2961&	\textbf{0.0283}       \  \textbf{0.2067} \\
                    Translation& 0.0105    \  0.0382& 0.0097   \  0.0327&	0.0081  \ 0.0360& 0.0012  \ 0.0135&	  \textbf{0.0009}  \  \textbf{0.0111} \\
					\hline  
		    \end{tabular}}
      }
       \label{tab:forward}
 }

\end{table*}
\begin{table*}[tbp]
 \scriptsize{
  
 	\caption{Ablation experiments on planar motion (degree)}
	\setlength{\tabcolsep}{1.5mm}{
			\scalebox{1.6}{
				\begin{tabular}{c c c c c c}
					\hline
					\multirow{2}{*}{\scriptsize{Seq.}}   
					& \scriptsize{\scriptsize{4pt-Lee}\cite{4pt_lee}}  
                       & \  \scriptsize{\scriptsize{4pt-Liu}\cite{4pt_liu}}  
                       & \ \scriptsize{\scriptsize{4pt-Sweeney}\cite{4pt_Sweeney}}
                       & \ \scriptsize{\scriptsize{Wu}\cite{Ding_mul}}  
                        & \scriptsize{\textbf{OURS}} \\
					\cline{2-6}
				   & \ \ ${\varepsilon _{\tiny{{\bf{R}}}}}$\quad ${\varepsilon _{{\bf{t}},{\rm{dir}}}}$      
                    & \ \ ${\varepsilon _{\tiny{{\bf{R}}}}}$ \quad ${\varepsilon _{{\bf{t}},{\rm{dir}}}}$      
                    & \ \ ${\varepsilon _{\tiny{{\bf{R}}}}}$ \quad ${\varepsilon _{{\bf{t}},{\rm{dir}}}}$      
                    & \ \ ${\varepsilon _{\tiny{{\bf{R}}}}}$ \quad ${\varepsilon _{{\bf{t}},{\rm{dir}}}}$    
                    & \ \ ${\varepsilon _{\tiny{{\bf{R}}}}}$ \quad ${\varepsilon _{{\bf{t}},{\rm{dir}}}}$  \\
					\hline
					Point&  0.4475  \  1.5211& 0.4095   \  1.5212&0.2001  \ 1.2556& 0.0452   \ 0.1108&   \textbf{0.0198} \  \textbf{0.0593} \\
					Pitch& 0.4118    \  0.8803& 0.4003   \  0.8602&	0.2437  \ 0.2937& 0.0549   \ 0.1153&	  \textbf{0.0251}  \  \textbf{0.0683} \\
					Roll& 0.5021    \  0.8605&	0.6037   \  1.0435&	0.2614  \ 0.3519& 0.0577  \ 0.1129&   \textbf{0.0247}  \  \textbf{0.0602} \\
					Rotation& 0.3370    \  0.9140& 0.5861   \  0.8430&	0.3671  \ 1.0095& 0.0565   \ 0.2652&	\textbf{0.0401}      \  \textbf{0.2160} \\
                    Translation& 0.0179    \  0.0347& 0.0277   \  0.0594&	0.0073  \ 0.0328& 0.0015  \ 0.0077&	  \textbf{0.0006}  \  \textbf{0.0063} \\
					\hline  
		    \end{tabular}}
      }
       \label{tab:planar}
      }
\end{table*}
 \begin{table*}[tbp]
 \scriptsize{
 	\caption{Ablation experiments on sideways motion (degree)}
	\setlength{\tabcolsep}{1.5mm}{
			\scalebox{1.6}{
				\begin{tabular}{c c c c c c}
					\hline
					\multirow{2}{*}{\scriptsize{Seq.}}   
					& \scriptsize{\scriptsize{4pt-Lee}\cite{4pt_lee}}  
                       & \  \scriptsize{\scriptsize{4pt-Liu}\cite{4pt_liu}}  
                       & \ \scriptsize{\scriptsize{4pt-Sweeney}\cite{4pt_Sweeney}}
                       & \ \scriptsize{\scriptsize{Wu}\cite{Ding_mul}}  
                        & \scriptsize{\textbf{OURS}} \\
					\cline{2-6}
				   & \ \ ${\varepsilon _{\tiny{{\bf{R}}}}}$\quad ${\varepsilon _{{\bf{t}},{\rm{dir}}}}$      
                    & \ \ ${\varepsilon _{\tiny{{\bf{R}}}}}$ \quad ${\varepsilon _{{\bf{t}},{\rm{dir}}}}$      
                    & \ \ ${\varepsilon _{\tiny{{\bf{R}}}}}$ \quad ${\varepsilon _{{\bf{t}},{\rm{dir}}}}$      
                    & \ \ ${\varepsilon _{\tiny{{\bf{R}}}}}$ \quad ${\varepsilon _{{\bf{t}},{\rm{dir}}}}$    
                    & \ \ ${\varepsilon _{\tiny{{\bf{R}}}}}$ \quad ${\varepsilon _{{\bf{t}},{\rm{dir}}}}$  \\
					\hline
					Point& 0.4852  \  1.5201& 0.4216   \  1.4121&0.3251  \ 1.2510& 0.0288   \ 0.0698&   \textbf{0.0153} \  \textbf{0.0523} \\
					Pitch& 0.3067    \  1.0145& 0.2883   \  0.9102&	0.3006  \ 0.4629& 0.0522   \ 0.0756&	  \textbf{0.0280}  \  \textbf{0.0489} \\
					Roll& 0.1184   \  1.0672&	0.1235   \  1.2905&	0.2346  \ 0.3823& 0.0779  \ 0.0864&   \textbf{0.0251}  \  \textbf{0.0532} \\
					Rotation& 0.7260    \  1.7610& 0.4931   \  1.3827&	0.6362  \ 2.0099& 0.0576   \ 0.2634&	 \textbf{0.0301}       \  \textbf{0.2072} \\
                    Translation& 0.0191    \  0.0527& 0.0148   \  0.0586&	0.0251  \ 0.0701& 0.0013  \ 0.0105&	  \textbf{0.0005}  \  \textbf{0.0084} \\
					\hline  
		    \end{tabular}}
      }
      \label{tab:sideways}
 }

\end{table*}

\begin{figure*}[tbp]
	\centering
	\includegraphics[width=0.9\linewidth]{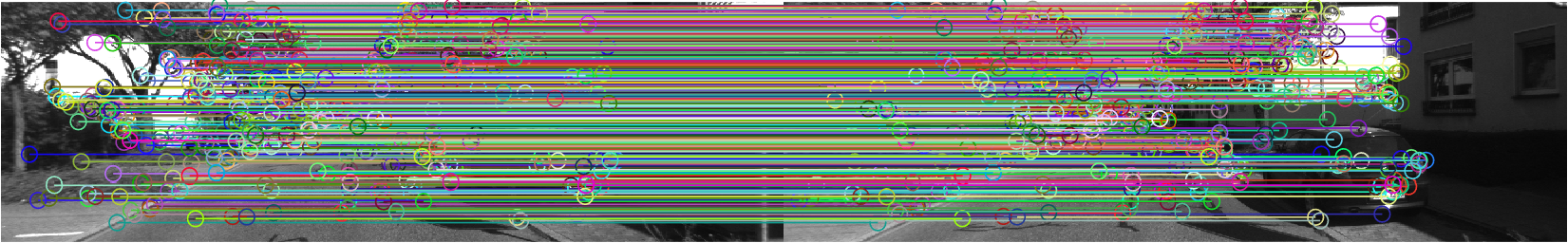}
	\caption{Test image pair from KITTI dataset with feature detection} 
	\label{fig.7}
\end{figure*}

\begin{figure*}[tbp]
	\centering 
	\includegraphics[width=1\linewidth]{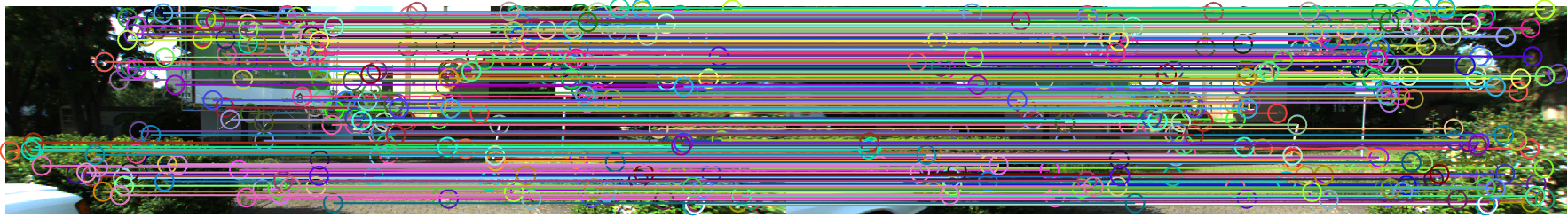}
	\caption{Test image pair from KITTI-360 dataset with feature detection} 
	\label{fig.kitti360}
\end{figure*}

\begin{figure}[tbp]             
  \centering
     \subfloat[Rotation error]{
     \begin{minipage}[t]{0.45\linewidth}
     \centering
     \includegraphics[width=0.96\linewidth]{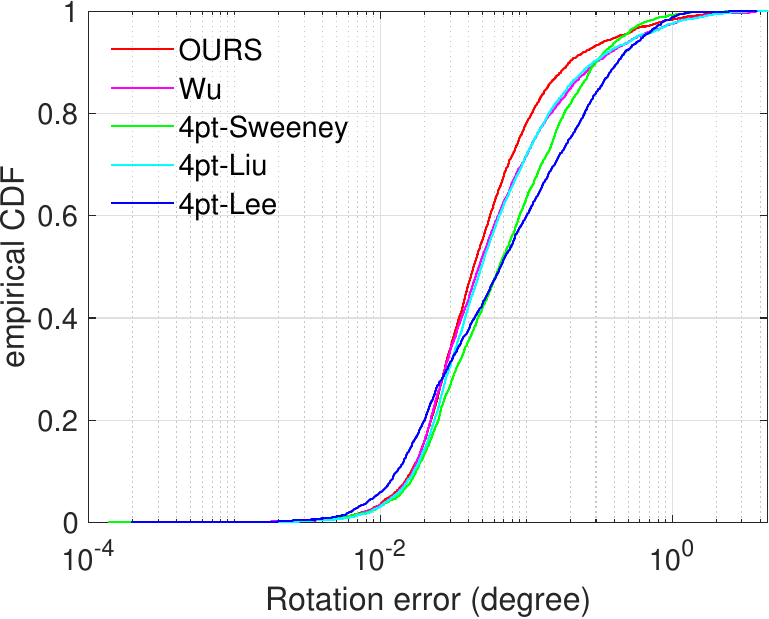}

     \end{minipage}
  
      }        
    \subfloat[Translation error]{
     \begin{minipage}[t]{0.45\linewidth}
     \centering
     \includegraphics[width=0.96\linewidth]{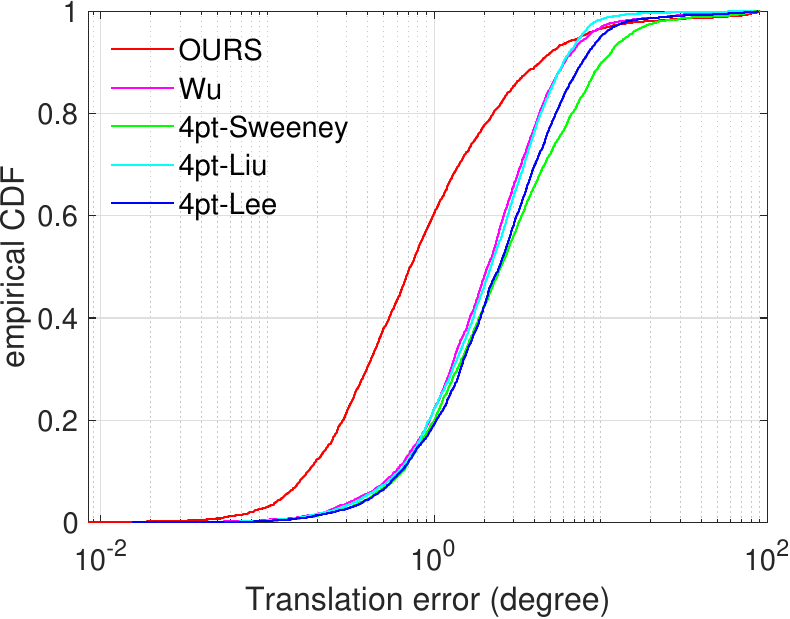}

     \end{minipage}
      } 
    \centering
    \caption{ Empirical cumulative error distributions for KITTI sequence 00.}
    \label{fig.6}
  \end{figure}

  \begin{figure}[tbp]             
  \centering
     \subfloat[Rotation error]{
     \begin{minipage}[t]{0.45\linewidth}
     \centering
     \includegraphics[width=0.96\linewidth]{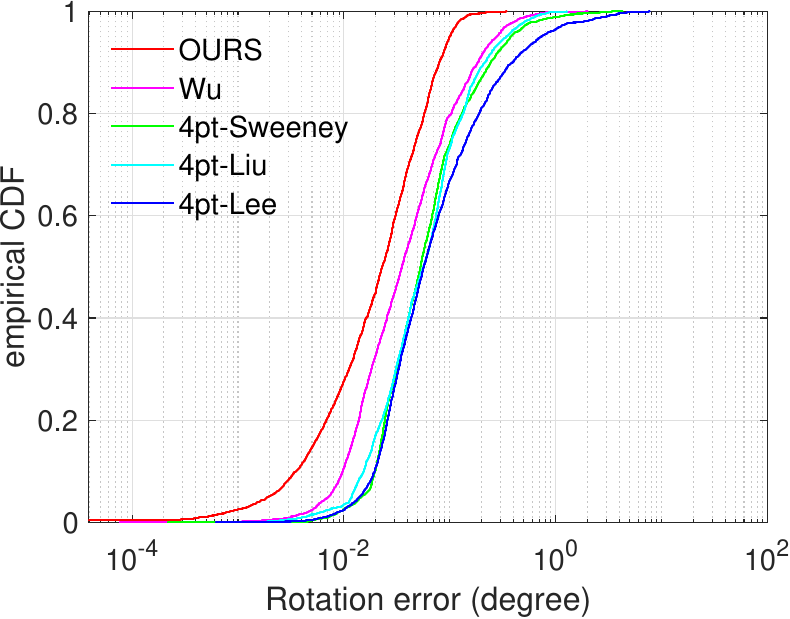}

     \end{minipage}
  
      }        
    \subfloat[Translation error]{
     \begin{minipage}[t]{0.45\linewidth}
     \centering
     \includegraphics[width=0.95\linewidth]{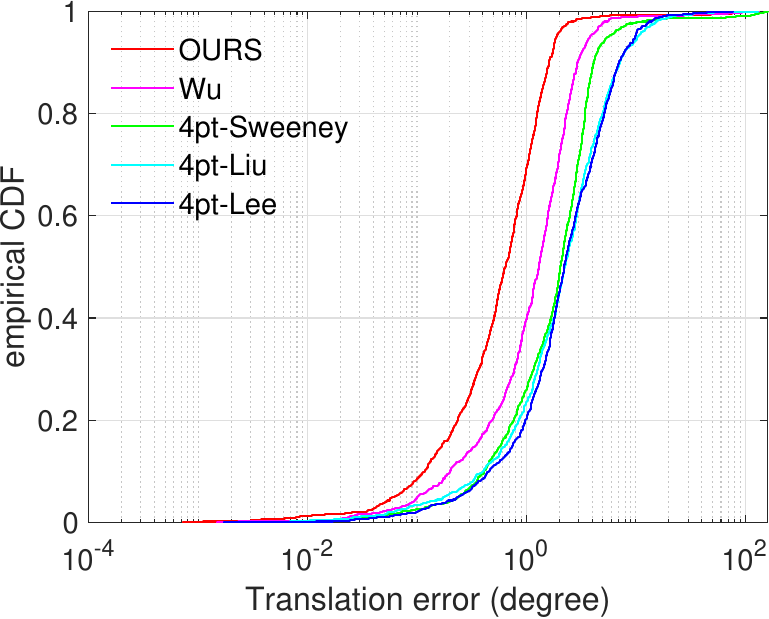}
   
     \end{minipage}
      } 
    \centering
    \caption{ Empirical cumulative error distributions for KITTI-360 sequence 00.}
    \label{fig.CDF360}
    \vspace{-20pt}
  \end{figure}

 \begin{table*}[tbp]
 \scriptsize{
  
 	\caption{Rotation and translation error for KITTI sequences with known vertical direction (degree)}
	\setlength{\tabcolsep}{1.5mm}{
			\scalebox{1.6}{
				\begin{tabular}{c c c c c c}
					\hline
					\multirow{2}{*}{\scriptsize{Seq.}}   
					& \scriptsize{\scriptsize{4pt-Lee}\cite{4pt_lee}}  
                       & \  \scriptsize{\scriptsize{4pt-Liu}\cite{4pt_liu}}  
                       & \ \scriptsize{\scriptsize{4pt-Sweeney}\cite{4pt_Sweeney}}
                       & \ \scriptsize{\scriptsize{Wu}\cite{Ding_mul}}  
                        & \scriptsize{\textbf{OURS}} \\
					\cline{2-6}
				   & \ \ ${\varepsilon _{\tiny{{\bf{R}}}}}$\quad \quad ${\varepsilon _{{\bf{t}},{\rm{dir}}}}$      
                    & \ \ ${\varepsilon _{\tiny{{\bf{R}}}}}$\quad \quad ${\varepsilon _{{\bf{t}},{\rm{dir}}}}$      
                    & \ \ ${\varepsilon _{\tiny{{\bf{R}}}}}$\quad \quad ${\varepsilon _{{\bf{t}},{\rm{dir}}}}$      
                    & \ \ ${\varepsilon _{\tiny{{\bf{R}}}}}$\quad \quad ${\varepsilon _{{\bf{t}},{\rm{dir}}}}$    
                    & \ \ ${\varepsilon _{\tiny{{\bf{R}}}}}$\quad \quad ${\varepsilon _{{\bf{t}},{\rm{dir}}}}$  \\
					\hline
					00& 0.065  \quad  \  2.469& 0.050  \quad \  2.190&	0.066 \quad \ 2.519& 0.049           \quad \ 2.089&   \textbf{0.043} \quad \  \textbf{0.719} \\
					01& 0.137  \quad  \  4.782& 0.115  \quad \  1.191&	0.105 \quad \ 3.781& 0.034           \quad \ 1.431&	  \textbf{0.023} \quad \  \textbf{0.212} \\
					02& 0.057  \quad  \  1.825&	0.044  \quad \  1.579&	0.057 \quad \ 1.975& 0.046           \quad \ 1.756&   \textbf{0.036} \quad \  \textbf{0.547} \\
					03& 0.064  \quad  \  3.116& 0.069  \quad \  3.712&	0.062 \quad \ 3.258& \textbf{0.045}  \quad \ 1.791&	     0.047       \quad \  \textbf{0.776} \\
                    04& 0.051  \quad  \  1.564& 0.045  \quad \  1.635&	0.051 \quad \ 1.708& 0.019           \quad \ 1.136&	  \textbf{0.016} \quad \  \textbf{0.631} \\
					05& 0.054  \quad  \  2.337& 0.052  \quad \  2.544&	0.056 \quad \ 2.406& 0.038           \quad \ 1.905&	  \textbf{0.027} \quad \  \textbf{0.524} \\
					06& 0.058  \quad  \  1.757& 0.092  \quad \  2.721&	0.056 \quad \ 1.760& 0.026           \quad \ 0.906&	  \textbf{0.024} \quad \  \textbf{0.362} \\
					07& 0.058  \quad  \  2.810& 0.065  \quad \  4.554&	0.054 \quad \ 3.048& 0.034           \quad \ 1.910&	  \textbf{0.026} \quad \  \textbf{0.652} \\
					08& 0.051  \quad  \  2.433& 0.046  \quad \  2.422&	0.053 \quad \ 2.457& 0.042           \quad \ 1.923&   \textbf{0.031} \quad \  \textbf{0.956} \\
					09& 0.056  \quad  \  1.838& 0.046  \quad \  1.656&	0.058 \quad \ 1.793& 0.037           \quad \ 1.273&   \textbf{0.034} \quad \  \textbf{0.650} \\
					10& 0.052 \quad   \  1.932& 0.040  \quad \  1.658&	0.058 \quad \ 1.888& 0.022           \quad \ 1.725&	  \textbf{0.013} \quad \  \textbf{0.823} \\  
					\hline  
		    \end{tabular}}
      }
      \label{tab:kitti} 
}

\end{table*}

 \begin{table*}[tbp]
 \scriptsize{
  
 	\caption{Rotation and translation error for KITTI-360 sequences with known vertical direction (degree)}
	\setlength{\tabcolsep}{1.5mm}{
			\scalebox{1.6}{
				\begin{tabular}{c c c c c c}
					\hline
					\multirow{2}{*}{\scriptsize{Seq.}}   
					& \scriptsize{\scriptsize{4pt-Lee}\cite{4pt_lee}}  
                       & \  \scriptsize{\scriptsize{4pt-Liu}\cite{4pt_liu}}  
                       & \ \scriptsize{\scriptsize{4pt-Sweeney}\cite{4pt_Sweeney}}
                       & \ \scriptsize{\scriptsize{Wu}\cite{Ding_mul}}  
                        & \scriptsize{\textbf{OURS}} \\
					\cline{2-6}
				   & \ \ ${\varepsilon _{\tiny{{\bf{R}}}}}$\quad \quad ${\varepsilon _{{\bf{t}},{\rm{dir}}}}$      
                    & \ \ ${\varepsilon _{\tiny{{\bf{R}}}}}$\quad \quad ${\varepsilon _{{\bf{t}},{\rm{dir}}}}$      
                    & \ \ ${\varepsilon _{\tiny{{\bf{R}}}}}$\quad \quad ${\varepsilon _{{\bf{t}},{\rm{dir}}}}$      
                    & \ \ ${\varepsilon _{\tiny{{\bf{R}}}}}$\quad \quad ${\varepsilon _{{\bf{t}},{\rm{dir}}}}$    
                    & \ \ ${\varepsilon _{\tiny{{\bf{R}}}}}$\quad \quad ${\varepsilon _{{\bf{t}},{\rm{dir}}}}$  \\
					\hline
					00& 0.050  \quad  \  2.371& 0.045  \quad \  2.579&	0.047 \quad \ 2.257& 0.037   \quad \ 1.725&   \textbf{0.025} \quad \  \textbf{0.768} \\
					02& 0.038 \quad  \  2.063&	0.034  \quad \  2.325&	0.037 \quad \ 2.285&0.028     \quad \ 1.505&   \textbf{0.026} \quad \  \textbf{0.430} \\
					03& 0.046  \quad  \  2.185& 0.051  \quad \  2.538&	0.056 \quad \ 2.359& 0.032   \quad \ 1.672&	    \textbf{0.029}      \quad \  \textbf{0.772} \\
                    04& 0.070  \quad  \  2.538& 0.089  \quad \  2.396&	0.081 \quad \ 2.516& 0.068  \quad \ 1.962&	  \textbf{0.059} \quad \  \textbf{0.856} \\
					05& 0.064  \quad  \  2.131& 0.067 \quad \  2.130&	0.068 \quad \ 2.120& 0.054      \quad \ 1.416&	  \textbf{0.049} \quad \  \textbf{0.985} \\
					06& 0.074 \quad  \  3.417& 0.062  \quad \  3.928&	0.067 \quad \ 3.305& 0.056          \quad \ 2.423&	  \textbf{0.051} \quad \  \textbf{0.938} \\
					07& 0.079  \quad  \  3.422& 0.072  \quad \  3.521&	0.073 \quad \ 3.210& 0.069           \quad \ 2.088&	  \textbf{0.046} \quad \  \textbf{1.098} \\
			      	09& 0.053  \quad  \  2.196& 0.058  \quad \  2.367&	0.060 \quad \ 2.637& \textbf{0.043}          \quad \ 1.523&   0.048 \quad \  \textbf{0.479} \\
					10& 0.057 \quad   \  2.215& 0.056  \quad \  2.233&	0.053 \quad \ 2.210& 0.035           \quad \ 1.295&	  \textbf{0.023} \quad \  \textbf{0.645} \\  
					\hline  
		    \end{tabular}}
      }
       \label{tab:kitti360} 
}

\end{table*}

\begin{figure}[htbp]             
  \centering
     \subfloat[4pt-Lee]{
     \begin{minipage}[t]{0.48\linewidth}
     \centering
     \includegraphics[width=0.9\linewidth]{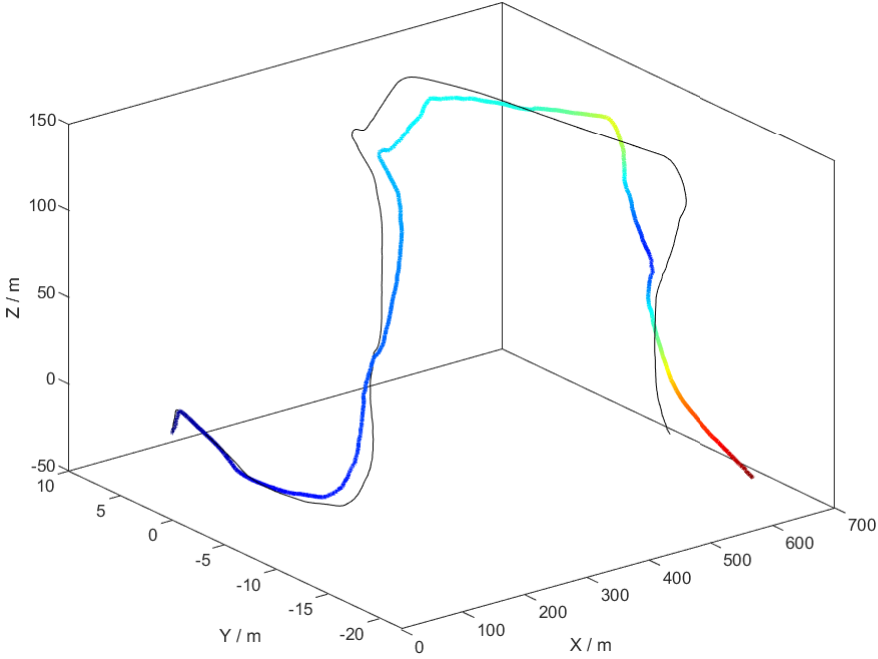}
     \label{fig.9a}
     \end{minipage}
      }
    \subfloat[4pt-Liu]{
     \begin{minipage}[t]{0.48\linewidth}
     \centering
     \includegraphics[width=0.9\linewidth]{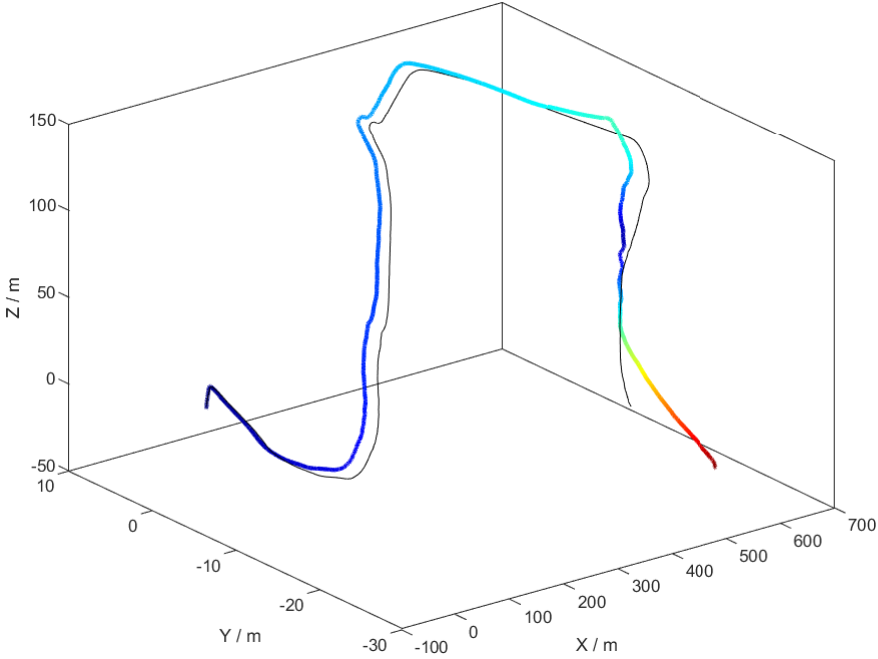}
    \label{fig.9b}
     \end{minipage}
      }

     \subfloat[4pt-Sweeney]{
     \begin{minipage}[t]{0.48\linewidth}
     \centering
     \includegraphics[width=0.9\linewidth]{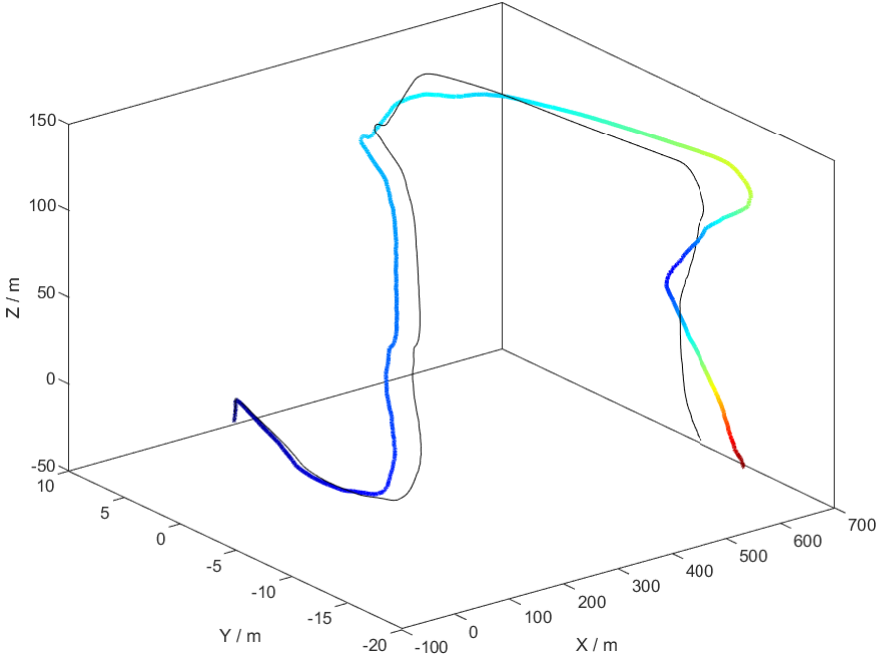}
     \label{fig.9c}
     \end{minipage}
      } 
      \subfloat[Wu]{
     \begin{minipage}[t]{0.48\linewidth}
     \centering
     \includegraphics[width=0.9\linewidth]{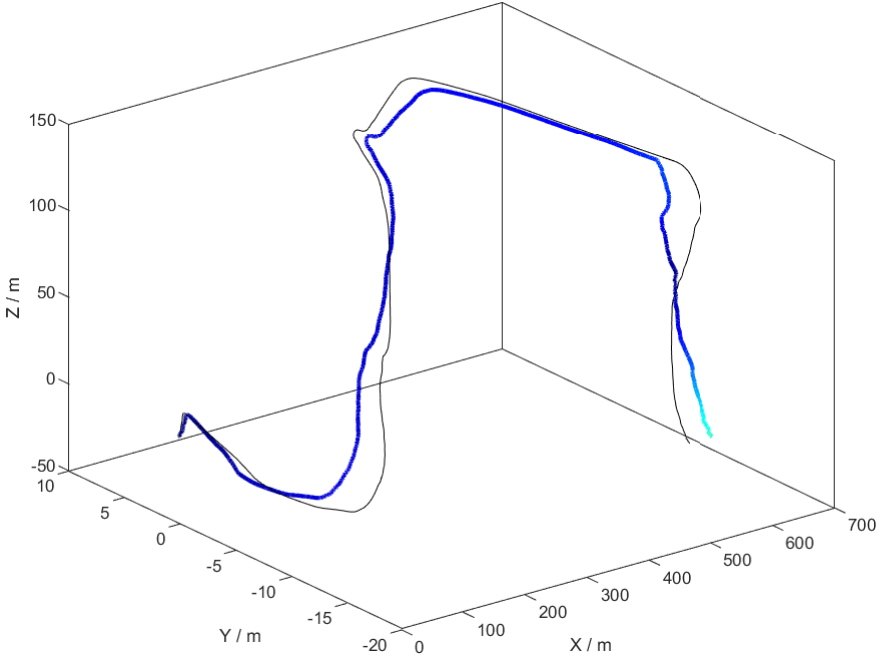}
     \label{fig.9d}
     \end{minipage}
      }

      \subfloat[OURS]{
     \begin{minipage}[t]{0.48\linewidth}
     \centering
     \includegraphics[width=0.9\linewidth]{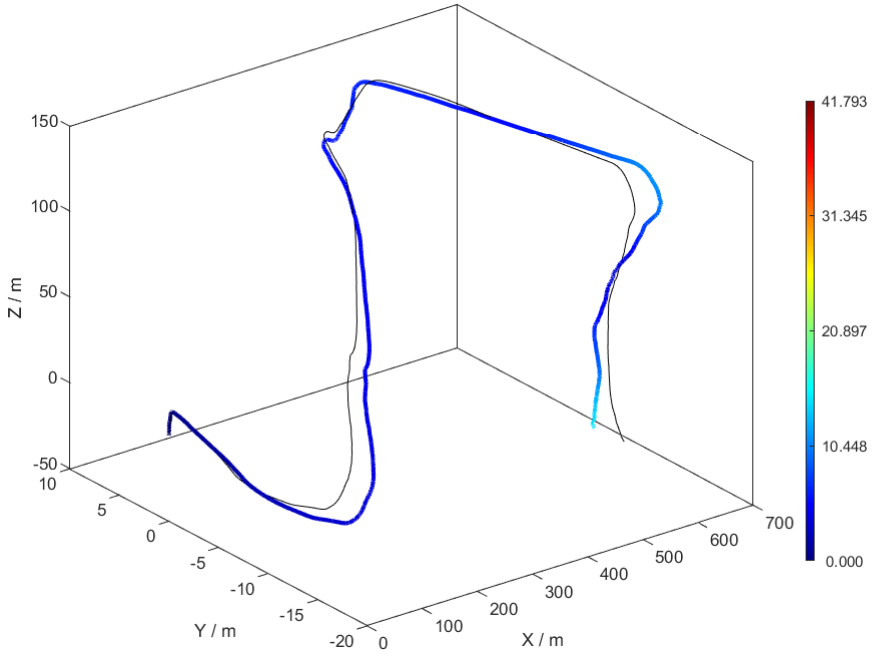}
     \label{fig.9e}
     \end{minipage}
      }       
   \vspace{-4pt}

    \centering
    \caption{Estimated trajectories plots for sequence 10 in the KITTI dataset. Black represents the ground truth trajectory. Colorful curves are estimated trajectories.}
    \label{fig.9}
   \vspace{-10pt}
\end{figure}

 \begin{figure}[htbp]             
  \centering
     \subfloat[4pt-Lee]{
     \begin{minipage}[t]{0.48\linewidth}
     \centering
     \includegraphics[width=0.9\linewidth]{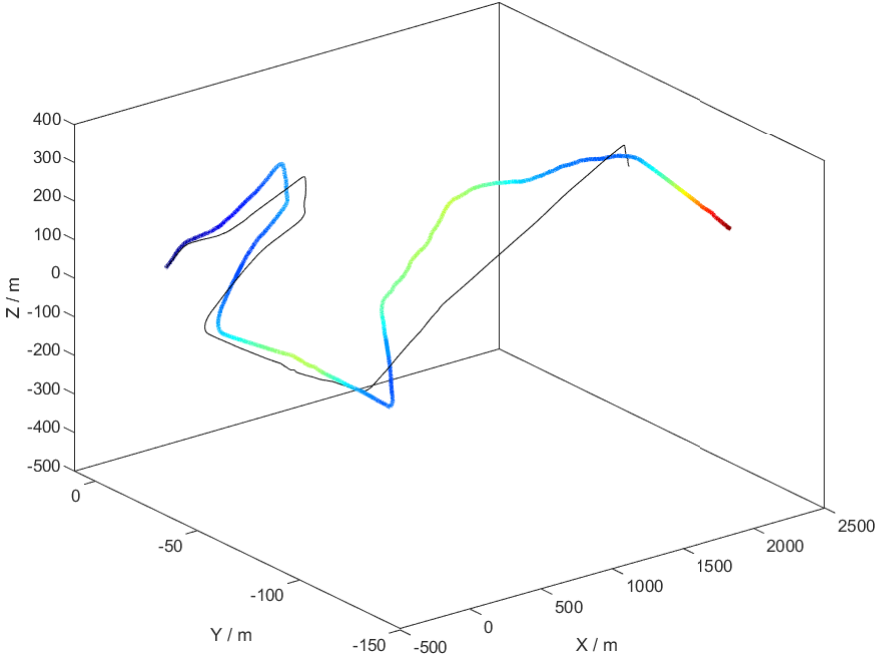}
     \label{fig.10a}
     \end{minipage}
      }
    \subfloat[4pt-Liu]{
     \begin{minipage}[t]{0.48\linewidth}
     \centering
     \includegraphics[width=0.9\linewidth]{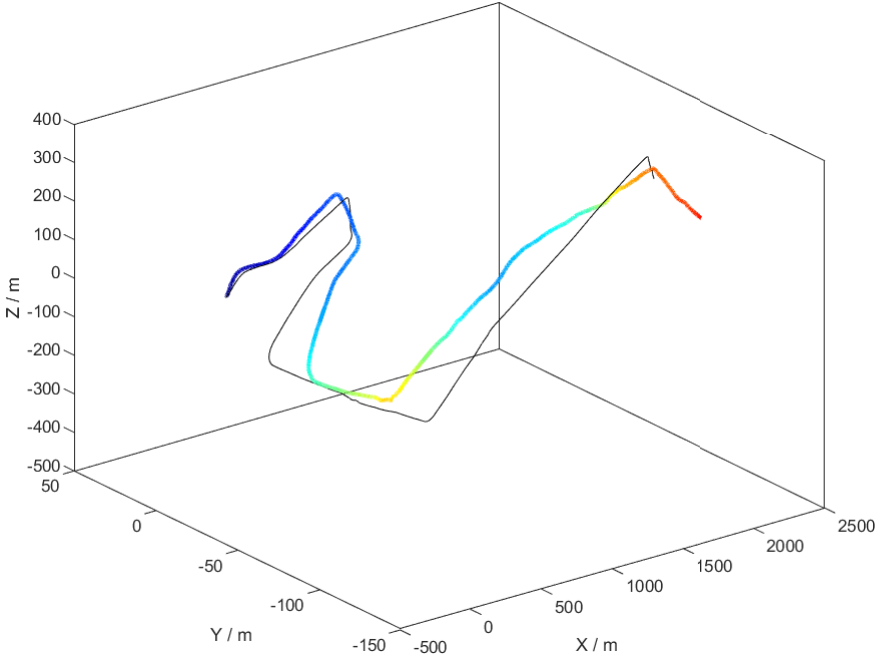}
    \label{fig.10b}
     \end{minipage}
      }

     \subfloat[4pt-Sweeney]{
     \begin{minipage}[t]{0.48\linewidth}
     \centering
     \includegraphics[width=0.9\linewidth]{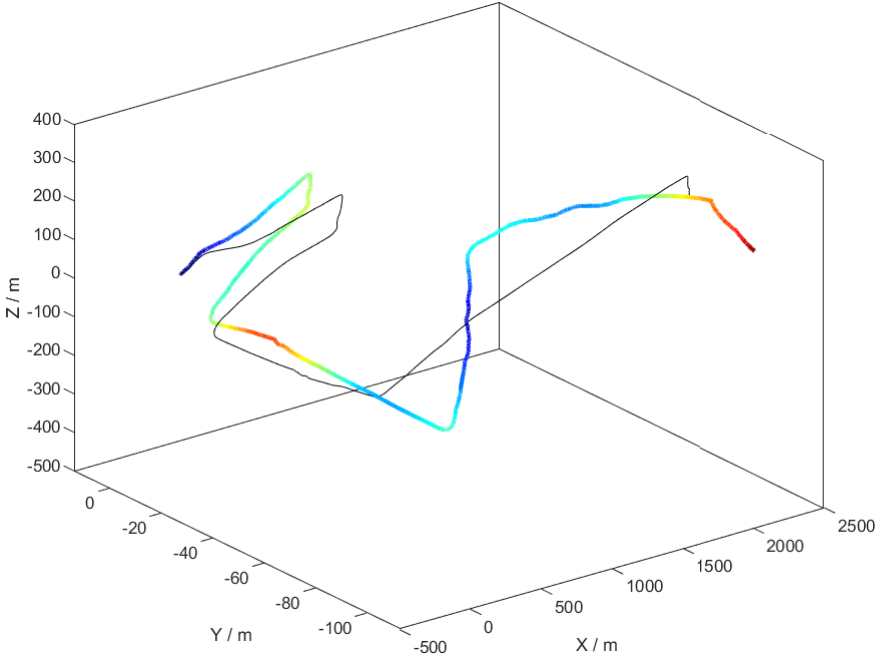}
     \label{fig.10c}
     \end{minipage}
      } 
      \subfloat[Wu]{
     \begin{minipage}[t]{0.48\linewidth}
     \centering
     \includegraphics[width=0.9\linewidth]{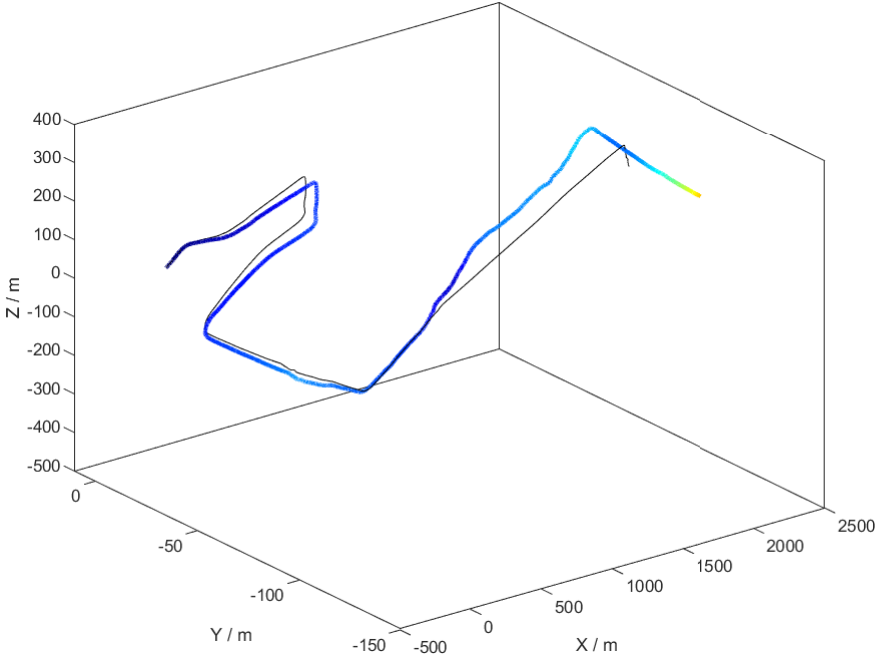}
     \label{fig.10d}
     \end{minipage}
      }

      \subfloat[OURS]{
     \begin{minipage}[t]{0.48\linewidth}
     \centering
     \includegraphics[width=0.9\linewidth]{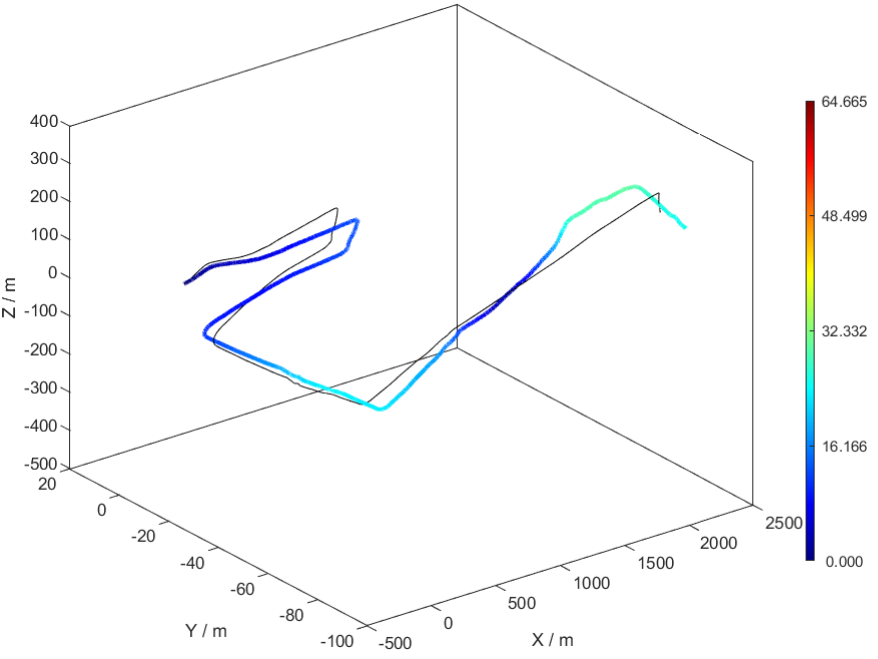}
     \label{fig.10e}
     \end{minipage}
      }       
   \vspace{-4pt}

    \centering
    \caption{Estimated trajectories plots for sequence 10 in the KITTI-360 dataset. Black represents the ground truth trajectory. Colorful curves are estimated trajectories.}
    \label{fig.10}
   \vspace{-10pt}
\end{figure}

\subsection{Ablation experiments}
According to Eq.~\eqref{24} and Eq.~\eqref{25}, we can easily observe that the error sources of the proposed method mainly include the accuracy of affine parameters, feature point, pitch angle, roll angle, rotation matrix, and translation vector. To analyze these factors, we conducted the following experiments: (1) One pixel noise is added to feature points, while pitch angle, roll angle, rotation matrix, and translation vector remain noise-free. (2) 0.2° noise is added to the pitch angle, with roll angle, feature points, rotation matrix, and translation vector being noise-free. (3) 0.2° noise is added to the roll angle, while the pitch angle, feature points, rotation matrix, and translation vector are noise-free. A perturbation of 0.01 is added to the rotation matrix, and the pitch angle, roll angle, feature points, and translation vector are noise-free. A perturbation of 0.01 is added to the translation vector, while pitch angle, roll angle, feature points, and rotation matrix are noise-free. Each method executes 1000 times.

Table~\ref{tab:random}, Table~\ref{tab:forward}, Table~\ref{tab:planar}, and Table~\ref{tab:sideways} show the ablation experiments conducted under random motion, forward motion, planar motion, and sideways motion, respectively. In these tables, bold font highlights the minimum value among the comparison methods. From Table~\ref{tab:random}-Table~\ref{tab:sideways}, it is evident that the rotation matrix and translation vector error estimated by the method proposed in this paper is an order of magnitude smaller compared to methods \texttt{4pt-Lee}, \texttt{4pt-Liu}, and \texttt{4pt-Sweeney}. Furthermore, the \texttt{OURS} method shows minimal sensitivity to noise under identical conditions across Table~\ref{tab:random}-Table~\ref{tab:sideways}. The proposed method demonstrates robust performance in accurately estimating multi-camera relative poses.

\subsection{Real data experiments}
To evaluate the accuracy of the proposed method in real-world scenarios, we choose the KITTI dataset, which is collected in an autonomous driving environment~\cite{KIttI}. The data acquisition platform is equipped with one inertial navigation system, one 64-line 3D LiDAR, and four cameras. The KITTI dataset is captured by driving around the mid-size city of Karlsruhe. It has ground-truth values for the relative pose of sequences 00-10. The vertical direction can be extracted from the IMU sensor. The intrinsic matrix of the cameras and the IMU data are provided in the data document~\cite{KIttI}. The KITTI-360 dataset is an extension of the KITTI dataset~\cite{liao2022kitti}. The KITTI-360 dataset records several suburbs of Karlsruhe, Germany. The car is equipped with a 180° fisheye camera on each side, and there are two 90° stereo perspective cameras in front of the car. We selecte images captured by these two perspective cameras to validate the performance of the method proposed in this paper. The system is also equipped with an IMU/GPS positioning system, providing ground truth. The intrinsic matrix of cameras, the IMU data, and pose data are provided in the data document~\cite{liao2022kitti}. It includes 9 sequences of rectified images. The affine correspondences in each camera between adjacent frames can be obtained by applying ASIFT~\cite{asift}. Fig.~\ref{fig.7} shows the results of feature extraction and matching using ASIFT in the KITTI dataset. Fig.~\ref{fig.kitti360} shows the results of feature extraction and matching using ASIFT in the KITTI-360 dataset. The \verb+Wu+ method is a global optimization method in a multi-camera system with a known vertical direction. We use the solver using minimum sample correspondences~\cite{IJCV} to calculate the inliers. Then, the \verb+OURs+ and \verb+Wu+ methods are used to solve the rotation and translation of the camera. To ensure the fairness of the experiment and further verify the superiority of the \verb+OURS+ method, we also choose the \verb+4pt-Lee+, \verb+4pt-Liu+ and \verb+4pt-Sweeney+ methods as comparison methods for the real data.

Table~\ref{tab:kitti} shows the rotation error and translation error calculated by the \verb+4pt-Lee+, \verb+4pt-Liu+, \verb+4pt-Sweeney+, \verb+Wu+ and \verb+OURS+ methods on the KITTI dataset. The bold value indicate the minimum error in each sequence. Both \verb+OURs+ and \verb+Wu+ are global optimization methods using non-minimum samples. We choose RANSAC~\cite{ransac} as a robust estimator using the \verb+4pt-Lee+, \verb+4pt-Liu+, and \verb+4pt-Sweeney+ methods to estimate the relative pose. The performances of the \verb+OURS+ and \verb+Wu+ methods are better than those of \verb+4pt-Lee+, \verb+4pt-Liu+, and \verb+4pt-Sweeney+ methods. Additionally, the \verb+OURS+ method outperforms the \verb+4pt-Lee+, \verb+4pt-Liu+, \verb+4pt-Sweeney+, and \verb+Wu+ methods, except for the rotation error of sequence 03. Fig.~\ref{fig.6} shows the empirical cumulative distribution function (CDF) of the rotation error and the translation error of sequence 00. This also shows the superiority of our method in estimating the generalized relative pose of multi-camera systems. Table~\ref{tab:kitti360} shows the error for the rotation matrix and translation vector estimation on the KITTI-360 dataset. The accuracy of the rotation matrix estimated by the \texttt{OURS} method is better than that of the \texttt{Wu}, \texttt{4pt-Sweeney}, \texttt{4pt-Liu}, and \texttt{4pt-Lee}, except for sequence 09. For the translation vector, the \texttt{OURS} method outperforms the other methods. Fig.~\ref{fig.CDF360} shows the empirical cumulative distribution function of the rotation error and the translation error for sequence 00 in the KITTI-360 dataset. The rotation error and translation vector error calculated by the \texttt{OURS} method are the smallest compared to the other methods shown in Table~\ref{tab:kitti360} and Fig.~\ref{fig.CDF360}. The superiority of the \texttt{OURS} method has been validated on both the KITTI and KITI-360 datasets.

To further demonstrate the effectiveness of the proposed method, we plotted the camera trajectories for comparison between different methods and evaluated them using Absolute Trajectory Error (ATE). Fig.~\ref{fig.9} shows the estimated trajectory plots for sequence 10 in the KITTI dataset. Fig.~\ref{fig.10} shows the estimated trajectory plots for sequence 10 in the KITTI-360 dataset. We directly plot trajectory graphs using computed relative rotation matrices and translation vectors between adjacent frames without any refinement. Black represents the ground truth trajectory, while the color coding of the trajectories indicates the Absolute Trajectory Error (ATE). Colorful curves are estimated trajectories. We can see that the method proposed in this paper outperforms the comparative methods.

\section{Conclusion}
We propose a novel globally optimal solver for generalized relative pose estimation. We utilize affine correspondences with non-minimum samples to solve the relative pose when the IMU and the camera are fixed. We establish the cost function by minimum algebraic error according to the least square criterion. Then, the cost function is transformed into two equations and two unknowns with the relative rotation angle. Then, the polynomial eigenvalue method is used to solve the parameters. Besides, considering the small rotation angle in practical application, a first-order approximate solver is proposed. The performance of our method is verified on simulated data and real data, which shows that the proposed method is more accurate than state-of-the-art methods.

According to Eq.~\eqref{25} and Eq.~\eqref{26}, it is evident that the main sources of error include the accuracy of feature points, the internal relationships within multi-camera systems (rotation matrix and translation vector), and the accuracy of affine parameters. The next task focuses on enhancing the calibration accuracy of multi-camera systems and improving feature extraction accuracy.

\bibliographystyle{IEEEtran}
\bibliography{reference}\ 

\begin{thebibliography}{10}
\providecommand{\url}[1]{#1}
\csname url@samestyle\endcsname
\providecommand{\newblock}{\relax}
\providecommand{\bibinfo}[2]{#2}
\providecommand{\BIBentrySTDinterwordspacing}{\spaceskip=0pt\relax}
\providecommand{\BIBentryALTinterwordstretchfactor}{4}
\providecommand{\BIBentryALTinterwordspacing}{\spaceskip=\fontdimen2\font plus
\BIBentryALTinterwordstretchfactor\fontdimen3\font minus
  \fontdimen4\font\relax}
\providecommand{\BIBforeignlanguage}[2]{{%
\expandafter\ifx\csname l@#1\endcsname\relax
\typeout{** WARNING: IEEEtran.bst: No hyphenation pattern has been}%
\typeout{** loaded for the language `#1'. Using the pattern for}%
\typeout{** the default language instead.}%
\else
\language=\csname l@#1\endcsname
\fi
#2}}
\providecommand{\BIBdecl}{\relax}
\BIBdecl

\bibitem{VO}
L.~Sv{\"a}rm, O.~Enqvist, F.~Kahl, and M.~Oskarsson, ``City-scale localization
  for cameras with known vertical direction,'' \emph{IEEE Trans. Pattern Anal.
  Mach. Intell.}, vol.~39, no.~7, pp. 1455--1461, 2016.

\bibitem{SLAM1}
R.~Mur-Artal and J.~D. Tard{\'o}s, ``Orb-slam2: An open-source slam system for
  monocular, stereo, and rgb-d cameras,'' \emph{IEEE Trans. Robot.}, vol.~33,
  no.~5, pp. 1255--1262, 2017.

\bibitem{SLAM2}
C.~Cadena, L.~Carlone, H.~Carrillo, Y.~Latif \emph{et~al.}, ``Past, present,
  and future of simultaneous localization and mapping: Toward the
  robust-perception age,'' \emph{IEEE Trans. Robot.}, vol.~32, no.~6, pp.
  1309--1332, 2016.

\bibitem{SLAM3}
L.~Heng, B.~Choi, Z.~Cui, M.~Geppert, S.~Hu \emph{et~al.}, ``Project
  autovision: Localization and 3d scene perception for an autonomous vehicle
  with a multi-camera system,'' in \emph{Proc. IEEE Int. Conf. Robot. Autom.},
  2019, pp. 4695--4702.

\bibitem{SMF1}
J.~L. Schonberger and J.~M. Frahm, ``Structure-from-motion revisited,'' in
  \emph{Proc. IEEE/CVF Conf. Comput. Vis. Pattern Recognit.}, 2016, pp.
  4104--4113.

\bibitem{SMF2}
M.~Pollefeys, D.~Nist{\'e}r \emph{et~al.}, ``Detailed real-time urban 3d
  reconstruction from video,'' \emph{Int. J. Comput. Vis.}, vol.~78, pp.
  143--167, 2008.

\bibitem{SMF3}
H.~Cui, X.~Gao, and S.~Shen, ``Mcsfm: Multi-camera-based incremental
  structure-from-motion,'' \emph{IEEE T IMAGE PROCESS.}, vol.~32, pp.
  6441--6456, 2023.

\bibitem{plucker}
R.~Pless, ``Using many cameras as one,'' in \emph{Proc. IEEE/CVF Conf. Comput.
  Vis. Pattern Recognit.}, vol.~2, 2003, pp. II--587.

\bibitem{6pt_Stewenius}
H.~Stew{\'e}nius, D.~Nist{\'e}r, M.~Oskarsson, and K.~Astr{\"o}m, ``Solutions
  to minimal generalized relative pose problems,'' in \emph{Workshop on
  omnidirectional vision}, vol.~1, no.~2.\hskip 1em plus 0.5em minus
  0.4em\relax Citeseer, 2005, p.~3.

\bibitem{4pt_lee}
G.~Hee~Lee, M.~Pollefeys, and F.~Fraundorfer, ``Relative pose estimation for a
  multi-camera system with known vertical direction,'' in \emph{Proc. IEEE/CVF
  Conf. Comput. Vis. Pattern Recognit.}, 2014, pp. 540--547.

\bibitem{4pt_liu}
L.~Liu, H.~Li, Y.~Dai, and Q.~Pan, ``Robust and efficient relative pose with a
  multi-camera system for autonomous driving in highly dynamic environments,''
  \emph{IEEE Trans. Intell Transp}, vol.~19, no.~8, pp. 2432--2444, 2018.

\bibitem{4pt_Sweeney}
C.~Sweeney, J.~Flynn, and M.~Turk, ``Solving for relative pose with a partially
  known rotation is a quadratic eigenvalue problem,'' in \emph{Proc. IEEE
  International Conference on 3D Vision}, 2014, pp. 483--490.

\bibitem{17pt_li}
H.~Li, R.~Hartley, and J.-h. Kim, ``A linear approach to motion estimation
  using generalized camera models,'' in \emph{Proc. IEEE/CVF Conf. Comput. Vis.
  Pattern Recognit.}, 2008, pp. 1--8.

\bibitem{guan_6DOF}
B.~Guan and J.~Zhao, ``Affine correspondences between multi-camera systems for
  6dof relative pose estimation,'' in \emph{Proc. Eur. Conf.
  Comput.Vision.}\hskip 1em plus 0.5em minus 0.4em\relax Springer, 2022, pp.
  634--650.

\bibitem{ICCV}
B.~Guan, J.~Zhao, D.~Barath, and F.~Fraundorfer, ``Minimal cases for computing
  the generalized relative pose using affine correspondences,'' in \emph{Proc.
  IEEE Int. Conf. Comput. Vis.}, 2021, pp. 6068--6077.

\bibitem{IJCV}
B.~Guan, J.~Zhao \emph{et~al.}, ``Minimal solvers for relative pose estimation
  of multi-camera systems using affine correspondences,'' \emph{Int. J. Comput.
  Vis.}, vol. 131, no.~1, pp. 324--345, 2023.

\bibitem{8pt_Kneip}
L.~Kneip and H.~Li, ``Efficient computation of relative pose for multi-camera
  systems,'' in \emph{Proc. IEEE/CVF Conf. Comput. Vis. Pattern Recognit.},
  2014, pp. 446--453.

\bibitem{zhao2020certifiably}
J.~Zhao, W.~Xu, and L.~Kneip, ``A certifiably globally optimal solution to
  generalized essential matrix estimation,'' in \emph{Proc. IEEE/CVF Conf.
  Comput. Vis. Pattern Recognit.}, 2020, pp. 12\,034--12\,043.

\bibitem{Ding_mul}
Q.~Wu, Y.~Ding, X.~Qi, J.~Xie, and J.~Yang, ``Globally optimal relative pose
  estimation for multi-camera systems with known gravity direction,'' in
  \emph{Proc. IEEE Int. Conf. Robot. Autom.}, 2022, pp. 2935--2941.

\bibitem{banglei2020relative}
B.~Guan, J.~Zhao, D.~Barath, and F.~Fraundorfer, ``Relative pose estimation for
  multi-camera systems from affine correspondences,'' \emph{arXiv:2306.12996},
  2020.

\bibitem{SOCP}
J.~H. Kim, R.~Hartley, J.-M. Frahm, and M.~Pollefeys, ``Visual odometry for
  non-overlapping views using second-order cone programming,'' in \emph{Proc.
  IEEE Asian Conference on Computer Vision}, 2007, pp. 353--362.

\bibitem{bound_bound}
J.~H. Kim, H.~Li, and R.~Hartley, ``Motion estimation for nonoverlapping
  multicamera rigs: Linear algebraic and ${L\infty}$ geometric solutions,''
  \emph{IEEE Trans. Pattern Anal. Mach. Intell.}, vol.~32, no.~6, pp.
  1044--1059, 2009.

\bibitem{5_Ding}
J.~Campos, J.~R. Cardoso, and P.~Miraldo, ``Poseamm: A unified framework for
  solving pose problems using an alternating minimization method,'' in
  \emph{Proc. IEEE Int. Conf. Robot. Autom.}\hskip 1em plus 0.5em minus
  0.4em\relax IEEE, 2019, pp. 3493--3499.

\bibitem{QCQP}
F.~Kahl and D.~Henrion, ``Globally optimal estimates for geometric
  reconstruction problems,'' \emph{Int. J. Comput. Vis.}, vol.~74, no.~1, pp.
  3--15, 2007.

\bibitem{bujnak20093d}
M.~Bujnak, Z.~Kukelova, and T.~Pajdla, ``3d reconstruction from image
  collections with a single known focal length,'' in \emph{Proc. IEEE Int.
  Conf. Comput. Vis.}\hskip 1em plus 0.5em minus 0.4em\relax IEEE, 2009, pp.
  1803--1810.

\bibitem{fitzgibbon2001simultaneous}
A.~W. Fitzgibbon, ``Simultaneous linear estimation of multiple view geometry
  and lens distortion,'' in \emph{Proc. IEEE/CVF Conf. Comput. Vis. Pattern
  Recognit.}, vol.~1.\hskip 1em plus 0.5em minus 0.4em\relax IEEE, 2001, pp.
  I--I.

\bibitem{ding2020efficient}
Y.~Ding, J.~Yang, and H.~Kong, ``An efficient solution to the relative pose
  estimation with a common direction,'' in \emph{Proc. IEEE Int. Conf. Robot.
  Autom.}\hskip 1em plus 0.5em minus 0.4em\relax IEEE, 2020, pp.
  11\,053--11\,059.

\bibitem{polynomial}
Z.~Kukelova, M.~Bujnak, and T.~Pajdla, ``Polynomial eigenvalue solutions to
  minimal problems in computer vision,'' \emph{IEEE Trans. Pattern Anal. Mach.
  Intell.}, vol.~34, no.~7, pp. 1381--1393, 2011.

\bibitem{Dosovitskiy2015FlowNetLO}
A.~Dosovitskiy, P.~Fischer, E.~Ilg, P.~H{\"a}usser, C.~Hazirbas, V.~Golkov,
  P.~van~der Smagt, D.~Cremers, and T.~Brox, ``Flownet: Learning optical flow
  with convolutional networks,'' \emph{Proc. IEEE Int. Conf. Comput. Vis.}, pp.
  2758--2766, 2015.

\bibitem{mayer2016large}
N.~Mayer, E.~Ilg, P.~Hausser, P.~Fischer, D.~Cremers, A.~Dosovitskiy, and
  T.~Brox, ``A large dataset to train convolutional networks for disparity,
  optical flow, and scene flow estimation,'' in \emph{Proc. IEEE/CVF Conf.
  Comput. Vis. Pattern Recognit.}, 2016, pp. 4040--4048.

\bibitem{vijayanarasimhan1704sfm}
S.~Vijayanarasimhan, S.~Ricco, C.~Schmid, R.~Sukthankar, and K.~Fragkiadaki,
  ``Sfm-net: Learning of structure and motion from video. arxiv 2017,''
  \emph{arXiv preprint arXiv:1704.07804}.

\bibitem{wang2021deep}
J.~Wang, Y.~Zhong, Y.~Dai, S.~Birchfield, K.~Zhang, N.~Smolyanskiy, and H.~Li,
  ``Deep two-view structure-from-motion revisited,'' in \emph{Proc. IEEE/CVF
  Conf. Comput. Vis. Pattern Recognit.}, 2021, pp. 8953--8962.

\bibitem{zhuang2021fusing}
B.~Zhuang and M.~Chandraker, ``Fusing the old with the new: Learning relative
  camera pose with geometry-guided uncertainty,'' in \emph{Proc. IEEE/CVF Conf.
  Comput. Vis. Pattern Recognit.}, 2021, pp. 32--42.

\bibitem{parameshwara2022diffposenet}
C.~M. Parameshwara, G.~Hari, C.~Ferm{\"u}ller, N.~J. Sanket, and Y.~Aloimonos,
  ``Diffposenet: Direct differentiable camera pose estimation,'' in \emph{Proc.
  IEEE/CVF Conf. Comput. Vis. Pattern Recognit.}, 2022, pp. 6845--6854.

\bibitem{xiao2022deepmle}
Y.~Xiao, L.~Li, X.~Li, and J.~Yao, ``Deepmle: A robust deep maximum likelihood
  estimator for two-view structure from motion,'' in \emph{Proc. IEEE
  International Conference on Intelligent Robots and Systems.}, 2022, pp.
  10\,643--10\,650.

\bibitem{AC_EX}
D.~Barath and L.~Hajder, ``Efficient recovery of essential matrix from two
  affine correspondences,'' \emph{IEEE Trans. Image Process.}, vol.~27, no.~11,
  pp. 5328--5337, 2018.

\bibitem{guan2020minimal}
B.~Guan, J.~Zhao, Z.~Li, F.~Sun, and F.~Fraundorfer, ``Minimal solutions for
  relative pose with a single affine correspondence,'' in \emph{Proc. IEEE/CVF
  Conf. Comput. Vis. Pattern Recognit.}, 2020, pp. 1929--1938.

\bibitem{guan2021relative}
B.~Guan, J.~Zhao, Z.~Li, F.~Sun, and F.~Friedrich, ``Relative pose estimation
  with a single affine correspondence,'' \emph{IEEE Trans. on Cybernetics},
  vol.~52, no.~10, pp. 10\,111--10\,122, 2021.

\bibitem{zhao_jiaocha}
J.~Zhao and B.~Guan, ``On relative pose recovery for multi-camera systems,''
  \emph{arXiv preprint arXiv:2102.11996}, 2021.

\bibitem{quan1999linear}
L.~Quan and Z.~Lan, ``Linear n-point camera pose determination,'' \emph{IEEE
  Trans. Pattern Anal. Mach. Intell.}, vol.~21, no.~8, pp. 774--780, 1999.

\bibitem{kneip2014opengv}
L.~Kneip and P.~Furgale, ``Opengv: A unified and generalized approach to
  real-time calibrated geometric vision,'' in \emph{Proc. IEEE Int. Conf.
  Robot. Autom.}, 2014, pp. 1--8.

\bibitem{ding2021globally}
Y.~Ding, D.~Barath, J.~Yang, H.~Kong, and Z.~Kukelova, ``Globally optimal
  relative pose estimation with gravity prior,'' in \emph{Proc. IEEE/CVF Conf.
  Comput. Vis. Pattern Recognit.}, 2021, pp. 394--403.

\bibitem{barath2019homography}
D.~Barath and Z.~Kukelova, ``Homography from two orientation-and
  scale-covariant features,'' in \emph{Proc. IEEE Int. Conf. Comput. Vis.},
  2019, pp. 1091--1099.

\bibitem{KIttI}
A.~Geiger, P.~Lenz, C.~Stiller, and R.~Urtasun, ``Vision meets robotics: The
  kitti dataset,'' \emph{Int. J. Robort Res.}, vol.~32, no.~11, pp. 1231--1237,
  2013.

\bibitem{liao2022kitti}
Y.~Liao, J.~Xie, and A.~Geiger, ``Kitti-360: A novel dataset and benchmarks for
  urban scene understanding in 2d and 3d,'' \emph{IEEE Trans. Pattern Anal.
  Mach. Intell.}, vol.~45, no.~3, pp. 3292--3310, 2022.

\bibitem{asift}
J.~M. Morel and G.~Yu, ``Asift: A new framework for fully affine invariant
  image comparison,'' \emph{SIAM J. Imag. Sci.}, vol.~2, no.~2, pp. 438--469,
  2009.

\bibitem{ransac}
M.~A. Fishler, ``Random sample consensus: A paradigm for model fitting with
  applications to image analysis and automated cartography,'' \emph{Commun,
  ACM}, vol.~24, no.~6, pp. 381--395, 1981.

\end{thebibliography}

\end{document}